\documentclass[sigconf,nonacm]{acmart}

\usepackage{xcolor}         %
\usepackage{multirow}
\usepackage{makecell}
\usepackage{comment}
\usepackage{wrapfig}
\usepackage{amsmath}
\usepackage[inline]{enumitem}
\usepackage{arydshln}
\usepackage[most]{tcolorbox}
\usepackage{subcaption}
\usepackage{graphicx}
\usepackage{hyperref} %
\usepackage{float}
\usepackage{lipsum}
\usepackage{adjustbox}

\newcommand{\eg}{\textit{e}.\textit{g}.}

\definecolor{mygreen}{RGB}{0, 0, 0}
\definecolor{cellgreen}{RGB}{255, 255, 255}
\definecolor{cellyellow}{HTML}{FFEB9C}
\definecolor{cellred}{HTML}{FFC7CE}

\AtBeginDocument{%
  }

\setcopyright{rightsretained}
\copyrightyear{2024}
\acmYear{2024}
\acmDOI{XXXXXXX.XXXXXXX}

\acmConference[AIBS '24]{The First Workshop on AI Behavioral Science}{August 25, 2024}{Barcelona, Spain}

\begin{document}

\title{How Different AI Chatbots Behave? Benchmarking Large Language Models in Behavioral Economics Games}

\author{Yutong Xie}
\email{yutxie@umich.edu}
\affiliation{%
  \institution{University of Michigan}
  \city{Ann Arbor}
  \state{Michigan}
  \country{USA}
}

\author{Yiyao Liu}
\authornote{These authors contributed equally to this research. }
\email{yiyaoliu@umich.edu}
\affiliation{%
  \institution{University of Michigan}
  \city{Ann Arbor}
  \state{Michigan}
  \country{USA}
}

\author{Zhuang Ma}
\authornotemark[1]
\email{davidmaz@umich.edu}
\affiliation{%
  \institution{University of Michigan}
  \city{Ann Arbor}
  \state{Michigan}
  \country{USA}
}

\author{Lin Shi}
\authornotemark[1]
\email{linshia@umich.edu}
\affiliation{%
  \institution{University of Michigan}
  \city{Ann Arbor}
  \state{Michigan}
  \country{USA}
}

\author{Xiyuan Wang}
\authornotemark[1]
\email{denniswx@umich.edu}
\affiliation{%
  \institution{University of Michigan}
  \city{Ann Arbor}
  \state{Michigan}
  \country{USA}
}

\author{Walter Yuan}
\email{walter.yuan@moblab.com}
\affiliation{%
  \institution{MobLab}
  \city{Pasadena}
  \state{California}
  \country{USA}
}

\author{Matthew O. Jackson}
\email{jacksonm@stanford.edu}
\affiliation{%
  \institution{Stanford University}
  \city{Stanford}
  \state{California}
  \country{USA}
}

\author{Qiaozhu Mei}
\email{qmei@umich.com}
\affiliation{%
  \institution{University of Michigan}
  \city{Ann Arbor}
  \state{Michigan}
  \country{USA}
}

\renewcommand{\shortauthors}{Xie et al.}

\begin{abstract}
The deployment of large language models (LLMs) in diverse applications requires a thorough understanding of their decision-making strategies and behavioral patterns. As a supplement to a recent study on the behavioral Turing test \cite{mei2024turing}, 
this paper presents a comprehensive analysis of five leading LLM-based chatbot families as they navigate a series of behavioral economics games. 
By benchmarking these AI chatbots, we aim to uncover and document both common and distinct behavioral patterns across a range of scenarios. 
The findings provide valuable insights into the strategic preferences of each LLM, highlighting potential implications for their deployment in critical decision-making roles.
\end{abstract}

\keywords{AI, Chatbot, Behavioral Economics Games, Turing Test}

\maketitle

\section{Introduction}

In the rapidly advancing field of artificial intelligence, large language models (LLMs) are playing a transformative role in decision-making across diverse domains. These AI systems, capable of engaging in conversations, offering guidance, and tackling complex decisions, are becoming increasingly indispensable in scenarios requiring nuanced, human-like judgment \cite{girotra2023ideas,chen2023emergence,eloundou2023gpts,lee2023benefits,shackelford2023we,bubeck2023sparks}. Understanding the behavioral patterns and decision-making strategies of AI chatbots is therefore critical. Such insights not only help optimize their performance in specific applications but also enable better assessment of their reliability and predictability, particularly in contexts involving significant responsibilities.

One recent study conducted by \citet{mei2024turing}, has primarily focused on the behavior of OpenAI ChatGPT variations through a Turing test involving classic behavioral economics games. This study has revealed intricate details about ChatGPT's behavioral patterns and preferences in scenarios designed to test \emph{trust}, \emph{fairness}, \emph{risk aversion}, \emph{altruism}, \emph{cooperation}, and other traits. 
However, it remains unclear whether these findings are unique to ChatGPT or if they extend to other LLMs like Meta Llama, Google Gemini, Anthropic Claude, and Mistral models. These models, while being influential in the AI sphere, have not been extensively studied in similar contexts. 
Moreover, although some research has explored particular traits (\eg, trust dynamics) among different AI models \cite{xie2024can}, analyses covering other behavioral dimensions are still lacking, raising questions about the generalizability of these behavioral traits across various scenarios and models.

\begin{figure*}[t]
    \centering
    \vspace{-15pt}
    \includegraphics[width=\textwidth]{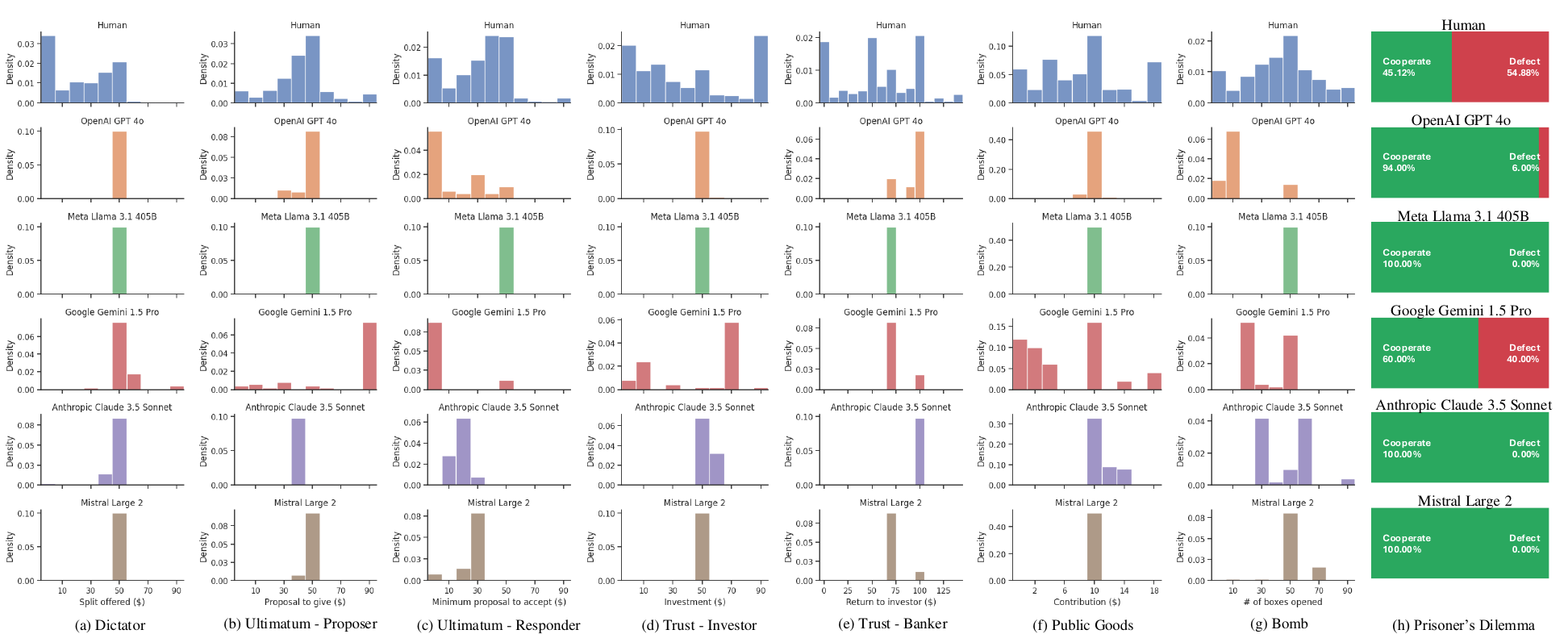}
    \vspace{-20pt}
    \caption{Distributions of AI chatbot behaviors in economics games. 
    }
    \vspace{-5pt}
    \label{fig:histograms}
\end{figure*}

As a supplementary work of \citet{mei2024turing}, this paper conducts a comprehensive analysis of five prominent LLM-based AI chatbot families through a series of behavioral economics games. By systematically evaluating their behaviors across these games, we aim to provide a detailed profile of these AI systems. Our study not only advances the understanding of AI behaviors but also highlights the nuanced differences that distinguish these models in decision-making contexts.
Some main findings are:

\begin{itemize}[leftmargin=15pt] 
    \item All tested chatbots successfully capture specific human behavior modes, leading to highly concentrated decision distributions (Fig. \ref{fig:histograms}).
    \item Although flagship chatbots demonstrate a notable probability of passing the Turing test (Fig. \ref{fig:turing}), AI chatbots can merely produce a behavior distribution similar to humans (Fig. \ref{fig:heatmap_w_dist_6_models}). 
    \item Compared to humans, AI chatbots place greater emphasis on maximizing fairness in their payoff preferences (Fig. \ref{fig:mse-avg}). 
    \item AI chatbots may exhibit inconsistencies in their payoff preferences across different games (Table \ref{tab:inconsistency}). 
    \item Different AI chatbots exhibit distinct behavioral patterns in games (Fig. \ref{fig:histograms}), which can be further distinguished through Turing test results (Fig. \ref{fig:turing}), revealed payoff preferences (Fig. \ref{fig:mse-avg}), and behavioral inconsistencies (Table \ref{tab:inconsistency}). These findings highlight the effectiveness of our behavioral benchmark in profiling and differentiating AI chatbots.
\end{itemize}

\section{Methods}

\subsection{LLM-Based AI Chatbots}

This study focuses on five families of LLM-based AI chatbots, as detailed in Table \ref{tab:models}. In the main text, results are presented exclusively for the flagship models. All model checkpoints were obtained as of July 31, 2024.

\begin{table}[h]
    \centering
    \vspace{-10pt}
    \begin{tabular}{|l|l|}
        \hline
        \textbf{AI chatbot family} & \textbf{Model variants/checkpoints} \\
        \hline
        OpenAI GPT
         & \texttt{gpt-4o-2024-05-13}$^{*}$ \\
         & \texttt{gpt-4o-mini-2024-07-18} \\
         & \texttt{gpt-4-0125-preview} \\
         & \texttt{gpt-4-0613} \\ 
         & \texttt{gpt-3.5-turbo-0125} \\
         & \texttt{gpt-3.5-turbo-0613} \\
         \hline
        Meta Llama
         & \texttt{llama-3.1-405B-instruct}$^{*}$ \\
         & \texttt{llama-3-70b-chat} \\
         & \texttt{llama-3-8b-chat} \\
         \hline
        Google Gemini 
         & \texttt{gemini-1.5-pro-latest}$^{*}$ \\
         & \texttt{gemini-1.0-pro-001} \\
         \hline
        Anthropic Claude 
         & \texttt{claude-3-5-sonnet-20240620}$^{*}$ \\
         & \texttt{claude-3-opus-20240229} \\
         & \texttt{claude-3-sonnet-20240229} \\
         & \texttt{claude-3-haiku-20240307} \\
         \hline
        Mistral 
        & \texttt{mistral-large-2407}$^{*}$ \\
         & \texttt{mistral-large-2402} \\
         
        \hline
    \end{tabular}
    \caption{LLM-based AI chatbots investigated in this study. In the main texts, we only report the results from flagship models as marked by ``$^*$''. 
    }
    \vspace{-30pt}
    \label{tab:models}
\end{table}

\subsection{Collecting AI Chatbot Behaviors in Economics Games}

Following \citet{mei2024turing}, we employ six classic behavioral economics games to evaluate multiple dimensions of AI behavior, including \emph{altruism}, \emph{fairness}, \emph{trust}, \emph{risk aversion}, and \emph{cooperation}. These games include Dictator, Ultimatum, Trust, Public Goods, Bomb Risk, and Prisoner’s Dilemma. Detailed descriptions of the games and the associated prompts can be found in \citet{mei2024turing}.

For each game and AI chatbot, we generate multiple responses using the respective game prompts, collecting 50 independent valid responses to establish the behavior distribution of each model. Human behavior distributions are taken from \citet{mei2024turing} for comparison.

\section{Results}

\begin{figure*}[htbp]
    \centering
    \vspace{-10pt}
    \includegraphics[width=0.9\textwidth]{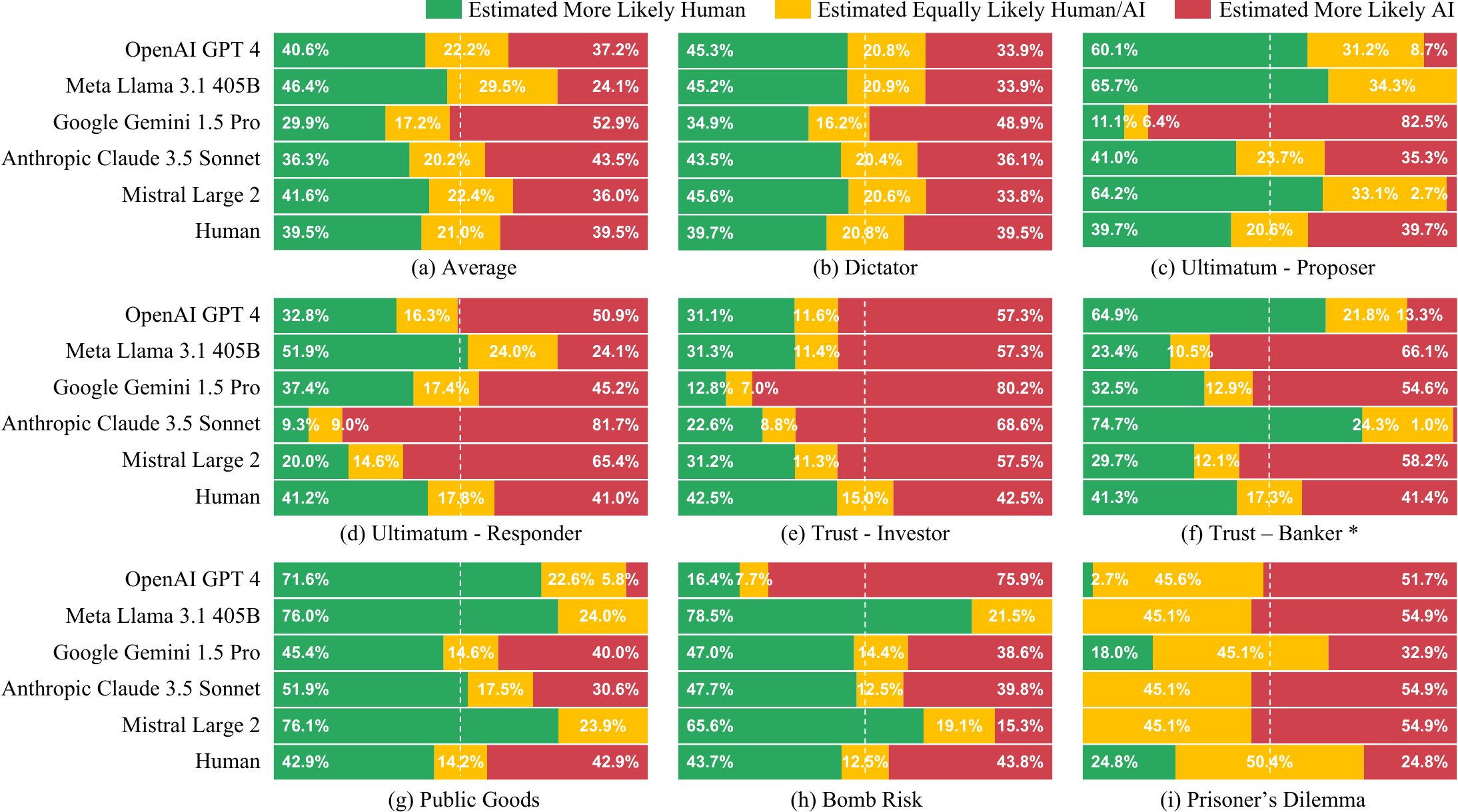}
    \vspace{-10pt}
    \caption{The Turing test results. }
    \vspace{-5pt}
    \label{fig:turing}
\end{figure*}

\begin{figure*}[htbp]  
    \centering
    \vspace{-5pt}
    \includegraphics[width=0.85\textwidth]{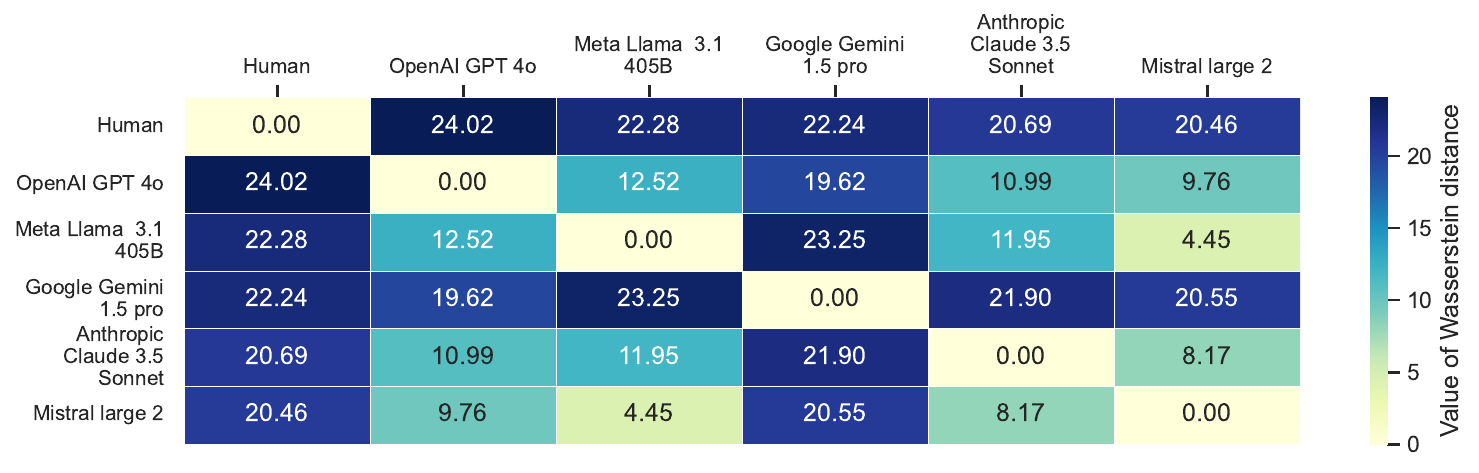}  
    \vspace{-15pt}
    \caption{Pairwise behavior distribution dissimilarities estimated with Wasserstein distance. 
    }
    \label{fig:heatmap_w_dist_6_models} 
\end{figure*}

\subsection{Behaviors of AI Chatbots}

Figure \ref{fig:histograms} (and Figure \ref{fig:variants} in the Appendix) illustrates the distributions of AI choices across the six games. Overall, the distributions of AI chatbots are notably more concentrated compared to human distributions, capturing only specific modes of human behavior. Additionally, different AI chatbots exhibit distinctly varied behavioral patterns, reflecting their unique orientations across multiple behavioral dimensions.

\vspace{-5pt}
\paragraph{Altruism. } In games including Dictator (Fig. \ref{fig:histograms}a) and Ultimatum - Proposer that reveal the altruism of players, AI chatbots display to be more altruistic than humans by offering more to the partner. Surprisingly, a large fraction of Google Gemini 1.5 Pro instances choose to offer most of the money (\$90-\$99) in Ultimatum - Proposer, showing its particularly high tendency of altruism. 

\vspace{-5pt}
\paragraph{Fairness. } Fairness is often emphasized by AI chatbots across games. 
In the Dictator (Fig. \ref{fig:histograms}a) and Ultimatum - Proposer Game (Fig. \ref{fig:histograms}b), most AI chatbots choose to offer \$50 to the partner, meaning a fair split. 
Correspondingly, Meta Llama 3.1 405B fairly requires a minimum split of \$50 as the Responder in Ultimatum (Fig. \ref{fig:histograms}c). 
Similarly in the Trust - Banker Game (Fig. \ref{fig:histograms}e), OpenAI GPT 4o and Anthropic Claude 3.5 Sonnet tend to return the investment and half the profit (\$100 in total) to the investor. 

\vspace{-5pt}
\paragraph{Trust. } The Trust Game (Fig. \ref{fig:histograms}d) particularly shows the trust dynamics. As the investor in the Trust investment game, AI chatbots possess different levels of trust towards the banker -- Anthropic Claude 3.5 Sonnet and Google Gemini 1.5 Pro display a higher trust level, investing \$53.20 and \$51.20 on average; While other models mostly invest \$50 to the banker. 

\vspace{-5pt}
\paragraph{Risk aversion. }
The Bomb Risk Game (Fig. \ref{fig:histograms}) tests the risk preference of players. In this game, Meta Llama 3.1 405B and Mistral Large 2 are risk-neural and choose to open 50 boxes most of the time, which yields the maximized expected payoff. However, compared to these two models, OpenAI GPT 4o and Google Gemini 1.5 Pro are more risk-averse, opening 14.06 and 35.46 boxes on average. Interestingly, Anthropic Claude 3.5 Sonnet displays different risk preferences across instances -- 44\% instances open less than 50 boxes, while 46\% instances open more than 50 boxes. 

\vspace{-5pt}
\paragraph{Cooperation. } In the Prisoner's Dilemma Game (Fig. \ref{fig:histograms}h), which shows the cooperation tendency of models, Meta Llama 3.1 405B, Anthropic Claude 3.5 Sonnet, and Mistral Large 2 have the highest proportion of choosing the Cooperation action (100\%). 
Google Gemini 1.5 Pro has the lowest rate of cooperating (60.00\%), but the ratio is still significantly larger than humans (45.12\%).
However, in the Public Goods Game (Fig. \ref{fig:histograms}f), most models tend to contribute half the money (\$10) to the group instead of a larger contribution. 

\subsection{The Behavioral Turing Test}

Using the collected behavior distributions of AI chatbots and the excerpted human behaviors, we conduct Turing tests following the methodology outlined in \citet{mei2024turing}. Adopting the same procedure as described in the paper, each round of the test involves independently sampling one human action and one action from the AI behavior distribution. These samples are then compared based on their probabilities within the human distribution.

Figure \ref{fig:turing} presents the results of the Turing tests. Overall, all tested AI chatbots demonstrate a remarkable ability to pass the Turing test, with Meta Llama 3.1 405B achieving the highest winning rate against humans at 46.4\%.

However, in certain games, the chatbots exhibit significant challenges in replicating human behavior. For instance, in the Trust Game - Investor role (Fig. \ref{fig:turing}e), AI chatbots tend to invest conservatively, whereas a substantial fraction of human players opt to invest their entire amount (Fig. \ref{fig:histograms}d). Similarly, in the Prisoner’s Dilemma (Fig. \ref{fig:turing}i), AI chatbots show a pronounced inclination toward cooperation, while over half of the human players choose to defect (Fig. \ref{fig:histograms}h).

AI chatbots exhibit diverse capabilities in passing the Turing test across different games. OpenAI GPT-4 shows a relatively high success rate in the Ultimatum - Proposer, Trust - Banker, and Public Goods games but struggles significantly in the Bomb Risk Game. Similarly, Meta Llama 3.1 405B performs well in many games but falls short in the Trust - Banker scenario. Anthropic Claude 3.5 Sonnet stands out in the Trust - Banker Game but barely passes the Turing test in the Ultimatum - Responder role.

\subsection{Behavior Distribution Similarity}

While the Turing test is a valuable method for evaluating an AI’s ability to act like a single human player \cite{turing1950computing}, it has inherent limitations in capturing the complete spectrum of the behavior distribution. To overcome these limitations, we introduce a complementary approach: a distribution similarity test that assesses whether AI chatbots can accurately represent the behavior distribution of a human population.

Table \ref{fig:heatmap_w_dist_6_models} (and Table \ref{fig:heatmap_w_dist_all_models} in the Appendix) presents the pairwise dissimilarities of behavior distributions, measured using the Wasserstein distance. Smaller distances indicate greater similarity between two distributions, whether comparing chatbots or humans and chatbots.

Among the AI chatbots, \texttt{gpt-3.5-turbo-0613} demonstrates the highest similarity to the human population (Fig. \ref{fig:heatmap_w_dist_all_models}), likely due to its ability to produce relatively diverse choices (Fig. \ref{fig:variants}). However, despite this similarity, a significant gap remains between the human behavior distribution and AI-generated actions, with no chatbot achieving a distribution that closely mirrors human behaviors.

We also observe relatively small Wasserstein distances among Meta Llama 3.1, Anthropic Claude models, and Mistral Large models (Fig. \ref{fig:heatmap_w_dist_all_models}), indicating that these chatbots exhibit similar behavioral patterns.

\subsection{Revealing the payoff Preferences}

To uncover the intrinsic objectives underlying the behaviors of AI chatbots, we perform analyses to identify and characterize their payoff preferences.

\paragraph{Objective optimization efficiency. }

\begin{figure}[htbp]
    \centering
    \begin{subfigure}[b]{\linewidth}
        \includegraphics[width=\linewidth]{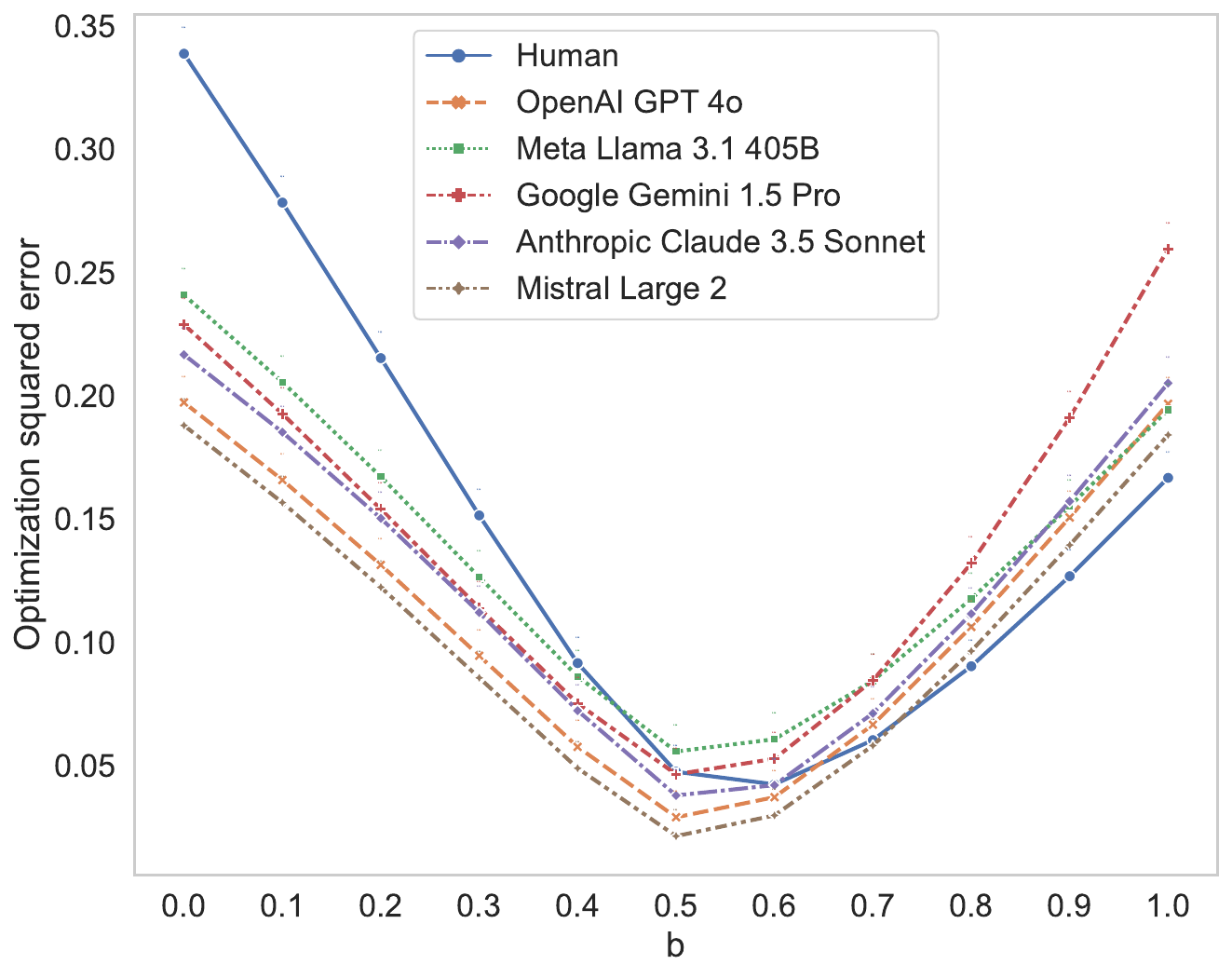}
        \vspace{-20pt}
        \caption{Linear specification ($r=1$). }
        \vspace{5pt}
    \end{subfigure}
    \begin{subfigure}[b]{\linewidth}
        \includegraphics[width=\linewidth]{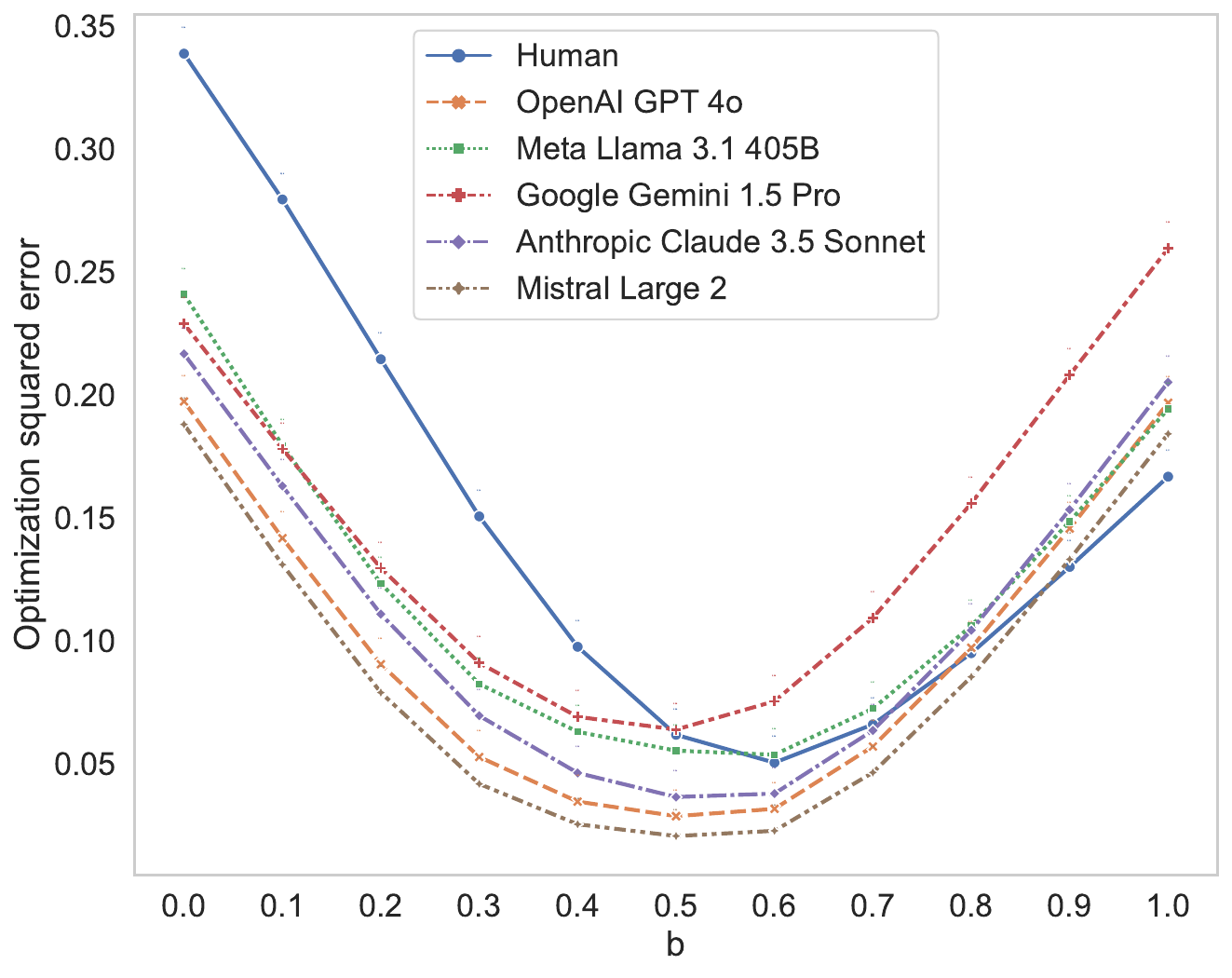}
        \vspace{-20pt}
        \caption{Non-linear (CES) specification ($r=1/2$). }
    \end{subfigure}
    \vspace{-20pt}
    \caption{
    The mean squared error (MSE) of the actual play distribution relative to the best-response utility, when matched with a partner playing the human distribution. The errors are calculated for each possible preference $b$ in the objective function (Eq. \ref{eq:utility}), and the average across all games is plotted. 
    Particularly,  $b = 1$  corresponds to purely selfish preferences,  $b = 0$  represents purely selfless preferences, and  $b = 0.5$  reflects a preference for maximizing the combined payoff of both players. $r$  is a specification parameter set as $r=1$ and $r=1/2$. 
    }
    \vspace{-10pt}
    \label{fig:mse-avg}
\end{figure}

The objective function of AI chatbots is quantitatively estimated by assessing the degree to which their behaviors align with the optimization goals. We adopt the family of utility functions from \citet{mei2024turing}:

\vspace{-10pt}
\begin{equation}
    \label{eq:utility}
    U_b=\left[b\times\text{Own payoff}^{\ r}+(1-b)\times\text{Partner payoff}^{\ r}\right]^{1/r},
\end{equation}

where  $b \in [0,1]$  represents the trade-off between a player’s own payoff and their partner’s payoff. Specifically,  $b = 1$  corresponds to purely selfish preferences,  $b = 0$  represents purely selfless preferences, and  $b = 0.5$  reflects a preference for maximizing the combined payoff of both players. In this context,  $r$  is a specification parameter that is frequently set to  $1$  (indicating a linear specification) or  $1/2$  (corresponding to a constant elasticity of substitution utility function, CES specification), as commonly adopted in the literature \cite{mcfadden1963constant}.

Figure \ref{fig:mse-avg} displays the optimization errors for various values of  $b$  for human players and each AI chatbot, computed under the assumption that the AI chatbots are interacting with a random human player. A lower optimization error indicates greater optimization efficiency, suggesting that the model is more likely to be optimizing for that particular objective.

The figure reveals that, in both utility specifications ($r = 1$  and  $r = 1/2$), AI chatbots place a stronger emphasis on fairness, as indicated by the lowest optimization error consistently occurring at $b = 0.5$. In contrast, human players exhibit a slight preference for selfishness, with their lowest optimization error occurring at $b = 0.6$.
Additionally, AI chatbots demonstrate significantly higher optimization efficiency than humans when maximizing the partner’s payoff ($b = 0$), but they exhibit lower efficiencies when optimizing their own payoff ($b = 1$).

Different AI models exhibit varying levels of optimization efficiency. For example, when optimizing for the partner’s payoff ($b=0$) or the combined payoff ( b=0.5 ), Mistral Large 2 achieves the highest optimization efficiency, followed by OpenAI GPT-4o. In contrast, Google Gemini 1.5 Pro and Meta Llama 3.1 405B show relatively lower optimization efficiency in these scenarios.
When optimizing for their own payoff ($b=1$), all chatbots perform with higher errors compared to humans, with Google Gemini 1.5 Pro displaying an even greater error than the other models.

Detailed optimization error plots for each individual game are provided in Figures \ref{fig:mse-diff-games-r1}–\ref{fig:mse-diff-games-r0.5} in the Appendix.

\paragraph{Logistic multinomial model fitting. }
In addition to the optimization efficiency analysis, a logistic multinomial model is also fitted to predict the behavior of AI chatbots. 
Following \citet{mei2024turing} , we assume AI takes action $k$ with a probability 

\vspace{-2pt}
\begin{equation}
    \text{Pr}(k)=\frac{\exp(U_b(k))}{\sum_{j\le K}\exp(U_b(j))},
\end{equation}
\vspace{2pt}

where $K$ is the number of all possible action choices. 

Table \ref{tab:beta-estimation1} summarizes the estimated  $b$  values based on the assumed logistic multinomial model. The table reveals that in nearly all games (except for the Ultimatum Game), the majority of AI chatbots exhibit estimated  $b$  values significantly lower than those of random human players. This indicates that, on average, AI chatbots tend to be more selfless compared to human players.

\subsection{Behavior Inconsistency}

\begin{table}[htbp]
    \centering
    \vspace{-5pt}
    \begin{tabular}{|l|c|c|}
        \hline
        \multirow{2}{*}{\textbf{Player}} & \multicolumn{2}{c|}{\textbf{Inconsistency}}  \\ \cline{2-3}
        & $r=1.0$ & $r=0.5$ \\ \hline
        Human players	&	0.114 & 0.122	\\ \hline
        OpenAI GPT 4o	&	0.115 & 0.107	\\ \hline
        Meta Llama 3.1 405B	&	0.125 & 0.125	\\ \hline
        Google Gemini 1.5 Pro	&	0.118 & 0.139	\\ \hline
        Anthropic Claude 3.5 Sonnet	&	0.143 & 0.133	\\ \hline
        Mistral Large 2	&	\textbf{0.108} & \textbf{0.100}	\\ 
        \hline
    \end{tabular}
    \caption{Behavior inconsistency across games of AI chatbots. The inconsistency is estimated by the mean absolute error of payoff curves (Fig. \ref{fig:mse-diff-games-r1}-\ref{fig:mse-diff-games-r0.5} in the Appendix). }
    \label{tab:inconsistency}
    \vspace{-20pt}
\end{table}

Although AI chatbots generally emphasize fairness and exhibit more selfless tendencies, they can display inconsistent behavior across different scenarios. For instance, a significant portion of Google Gemini 1.5 Pro instances choose to split the money fairly in the Dictator Game, yet in the Ultimatum - Proposer role, many instances propose offering nearly all the money (\$90-\$99) to the partner, reflecting an altruistic trait.

Table \ref{tab:inconsistency} (and Table \ref{tab:inconsistency-all} in the Appendix) provides the estimated behavior inconsistencies of AI chatbots. These inconsistencies are measured using the mean absolute error (MAE) of payoff curves across different games (see Figures \ref{fig:mse-diff-games-r1}–\ref{fig:mse-diff-games-r0.5} in the Appendix).

Among the flagship AI chatbots, Mistral Large 2 demonstrates the highest consistency across all games, as its own payoff and the partner's payoff are well-balanced in most scenarios (Fig. \ref{fig:histograms}). In contrast, Google Gemini 1.5 Pro and Anthropic Claude 3.5 Sonnet exhibit higher levels of inconsistency, even surpassing that of human players -- despite the fact that human behaviors reflect the diversity of a heterogeneous population.
\section{Discussion}

\paragraph{Model checkpoints. }
As AI chatbots evolve, their behavioral tendencies shift over time. Figures \ref{fig:variants}(i,ii) illustrate these changes across checkpoints for OpenAI GPT-4 and GPT-3 models.
For GPT-4, except for the Bomb Risk Game, the latest checkpoint (\texttt{gpt-4-0125-preview}) produces more concentrated behavior distributions compared to older checkpoints. The updated version also demonstrates higher rationality in the Ultimatum - Responder Game but shows increased risk aversion in the Bomb Risk Game.
For GPT-3, while the distribution modes largely remain consistent, the behavior distributions for the Ultimatum - Responder, Trust - Banker, and Bomb Risk games have become less diverse over successive updates.%

\paragraph{Model size. }
In addition to different checkpoints, variations in model size can also influence behavior. As shown in Figures \ref{fig:variants}(iii,iv), Meta Llama 3 8B behaves notably differently from the 70B version. In games like Ultimatum - Proposer, Trust - Investor, Public Goods, and Prisoner’s Dilemma, the 8B model exhibits more conservative tendencies.
For both Llama 3 and Anthropic Claude 3, smaller models (\textit{e.g.,} Llama 3 8B, Claude 3 Sonnet, and Claude 3 Haiku) display higher diversity in their behavior distributions compared to their larger counterparts.

\section{Conclusion and Future Work}

This study benchmarked LLM-based AI chatbots across a series of behavioral economics games. The analyses revealed the following common and distinct behavioral patterns of the chatbots:
\begin{enumerate*}
    \item All tested chatbots successfully capture specific human behavior modes, leading to highly concentrated decision distributions;
    \item Although flagship chatbots demonstrate a notable probability of passing the Turing test, AI chatbots can merely produce a behavior distribution similar to humans; 
    \item Compared to humans, AI chatbots place greater emphasis on maximizing fairness in their payoff preferences;
    \item AI chatbots may exhibit inconsistencies in their payoff preferences across different games;
    \item Different AI chatbots exhibit distinct behavioral patterns in games, which can be further distinguished in our analyses. 
\end{enumerate*}
These findings highlight the effectiveness of our behavioral benchmark in profiling and differentiating AI chatbots.

We hope our research contributes to a deeper understanding of AI behaviors and serves as a foundation for future studies in \emph{AI behavioral science}. For example, the discrepancies between Turing test results and distribution dissimilarities highlight the need for further alignment objectives that enable LLMs to better represent the diversity of the human behaviors. Additionally, the observed inconsistencies in AI behaviors across games underscore the importance of developing generalizable preferences and objectives for AI systems that can adapt effectively across various scenarios.

\newpage
\bibliographystyle{ACM-Reference-Format}
\bibliography{sample-base}

\appendix

\newpage
\section*{Appendix}

\begin{figure*}[h]  
    \centering
    \includegraphics[width=\textwidth]{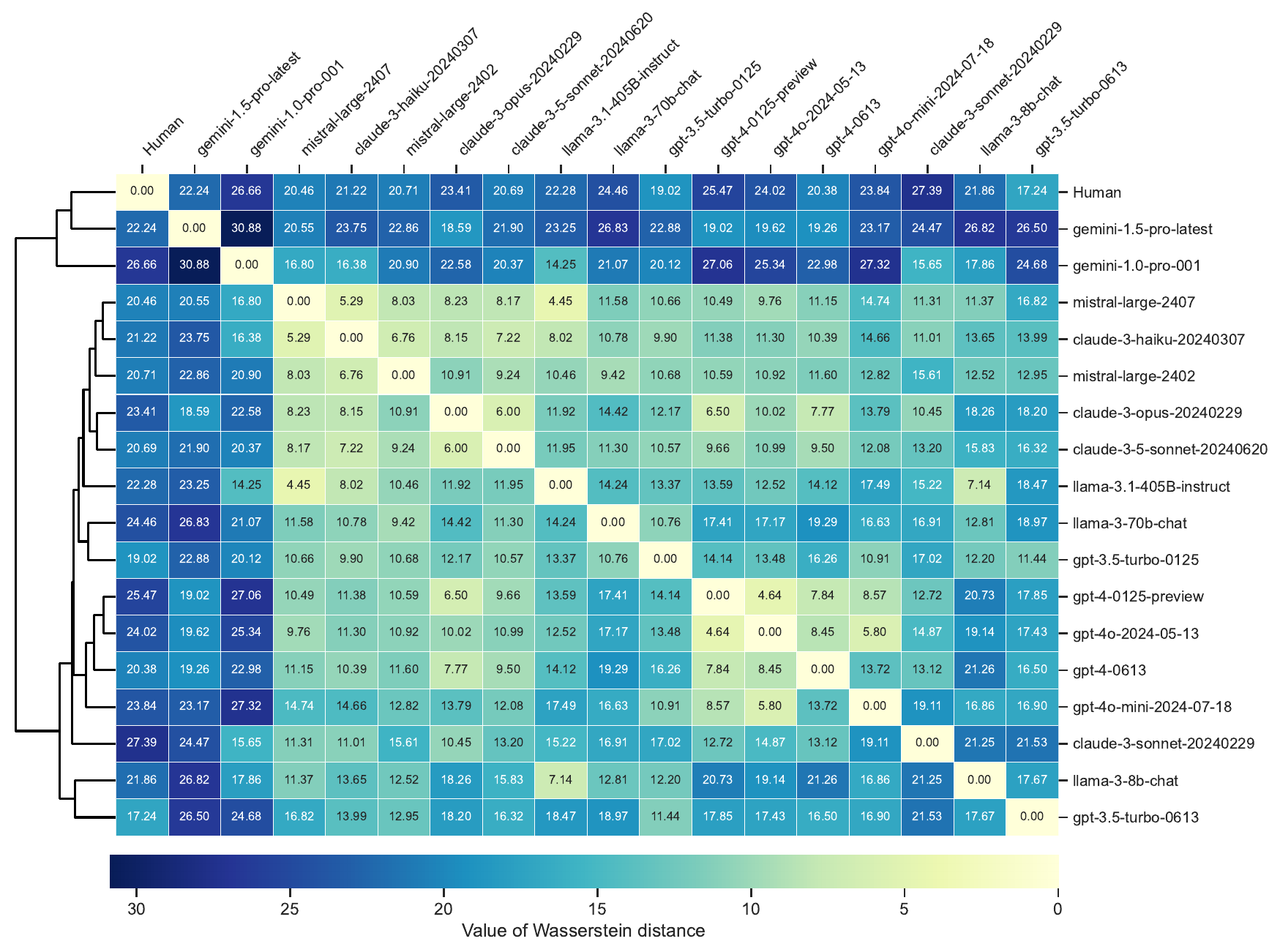}  
    \vspace{-15pt}
    \caption{Pairwise behavior distribution dissimilarities estimated with Wasserstein distance. 
    }
    \label{fig:heatmap_w_dist_all_models} 
\end{figure*}

\begin{figure*}[h]
    \centering
    \begin{subfigure}[b]{0.24\textwidth}
        \includegraphics[width=\textwidth]{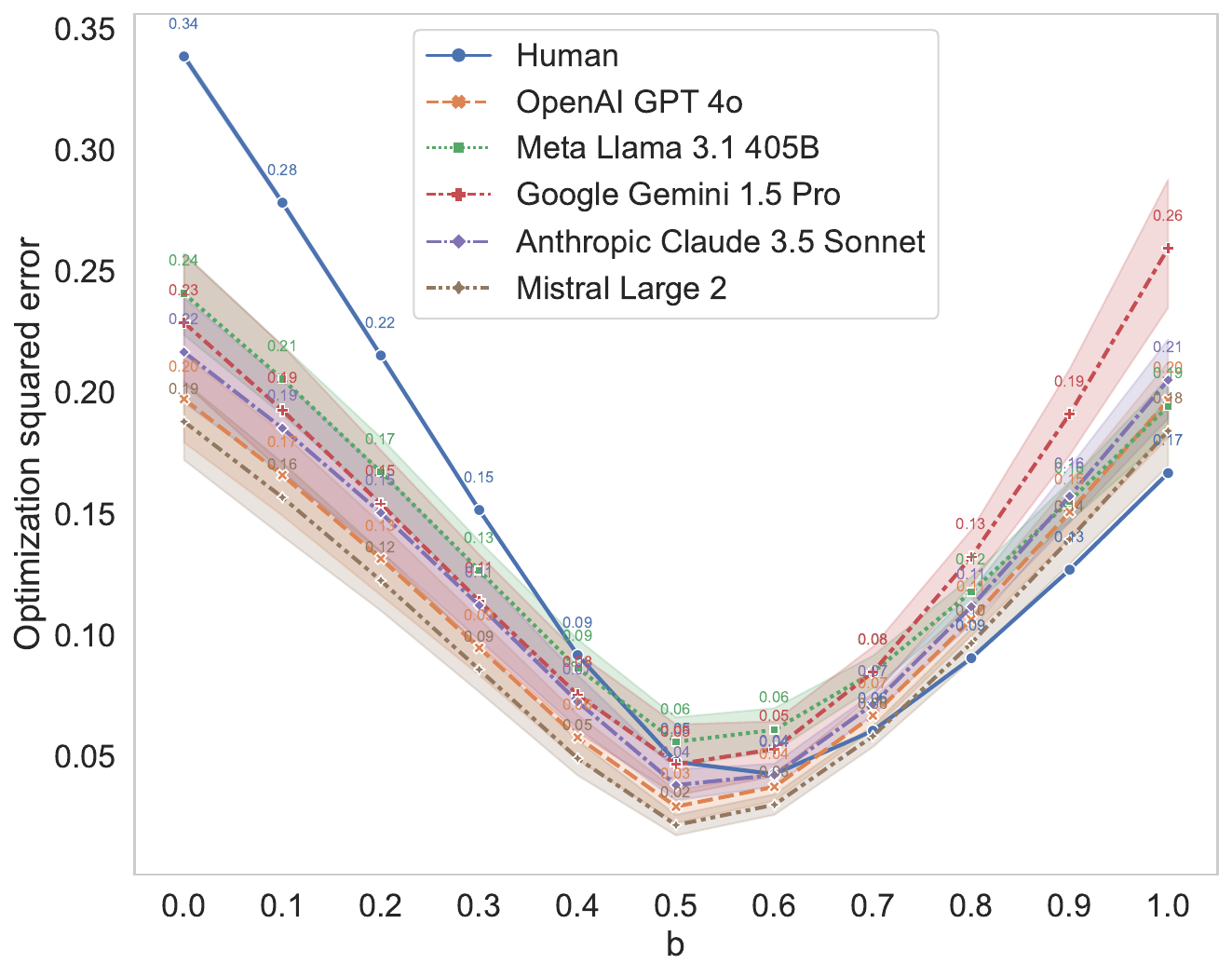}
        \caption{Average}
    \end{subfigure}
    \begin{subfigure}[b]{0.24\textwidth}
        \includegraphics[width=\textwidth]{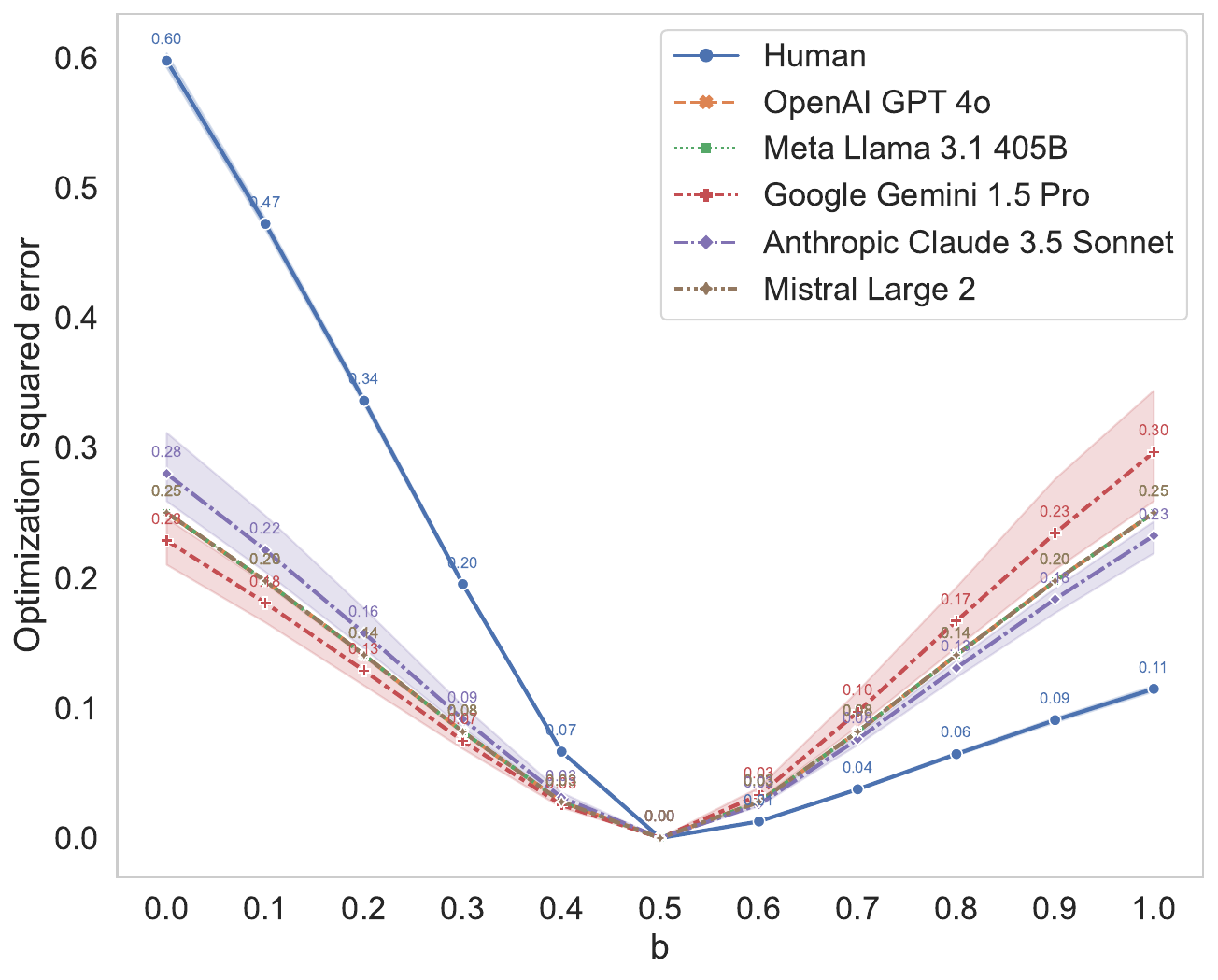}
        \caption{Dictator}
    \end{subfigure}
    \begin{subfigure}[b]{0.24\textwidth}
        \includegraphics[width=\textwidth]{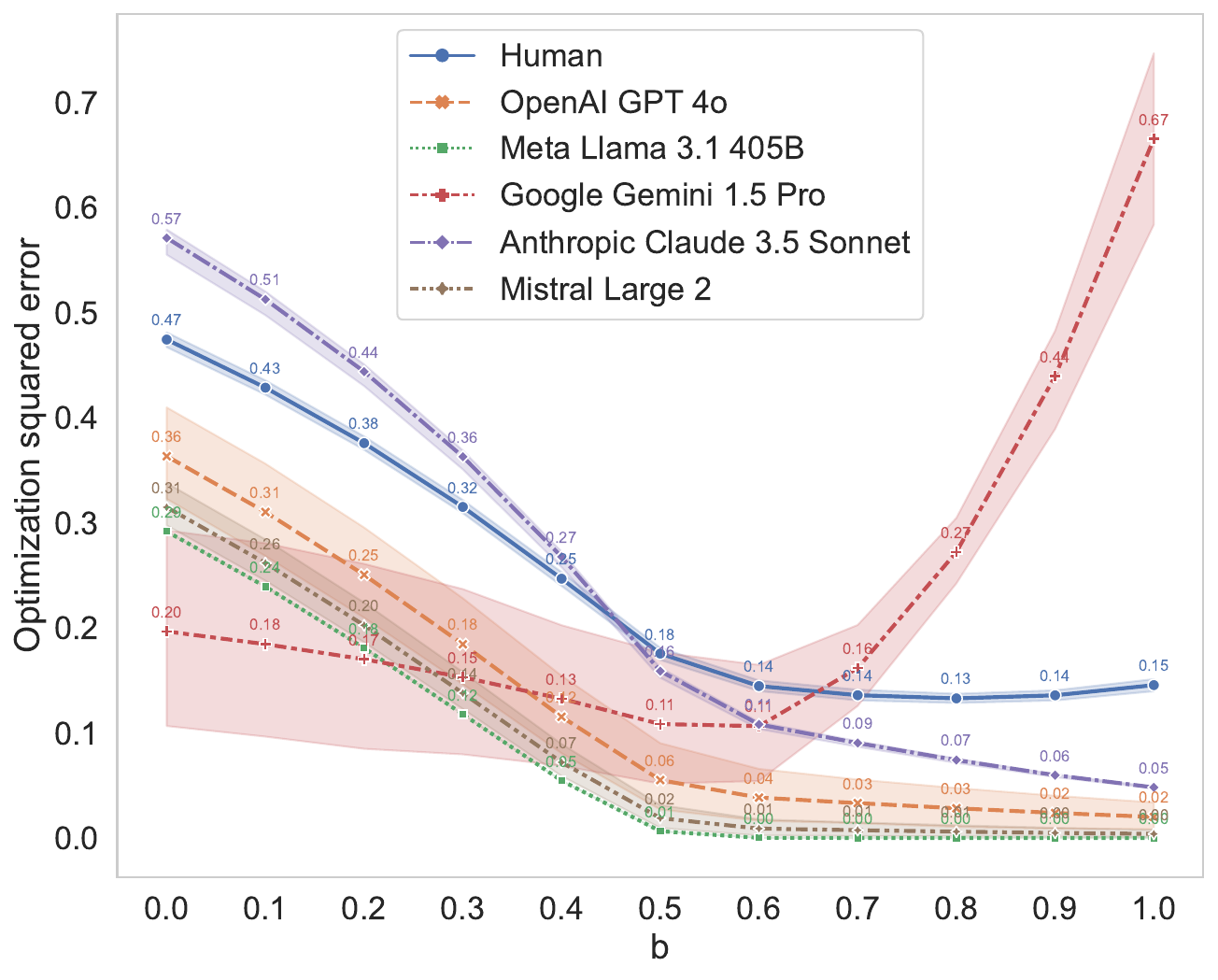}
        \caption{Ultimatum - Proposer}
    \end{subfigure}
    \begin{subfigure}[b]{0.24\textwidth}
        \includegraphics[width=\textwidth]{figures/fig7_8/ultimatum_1_1_new.pdf}
        \caption{Ultimatum - Responder}
    \end{subfigure}
    
    \begin{subfigure}[b]{0.24\textwidth}
        \includegraphics[width=\textwidth]{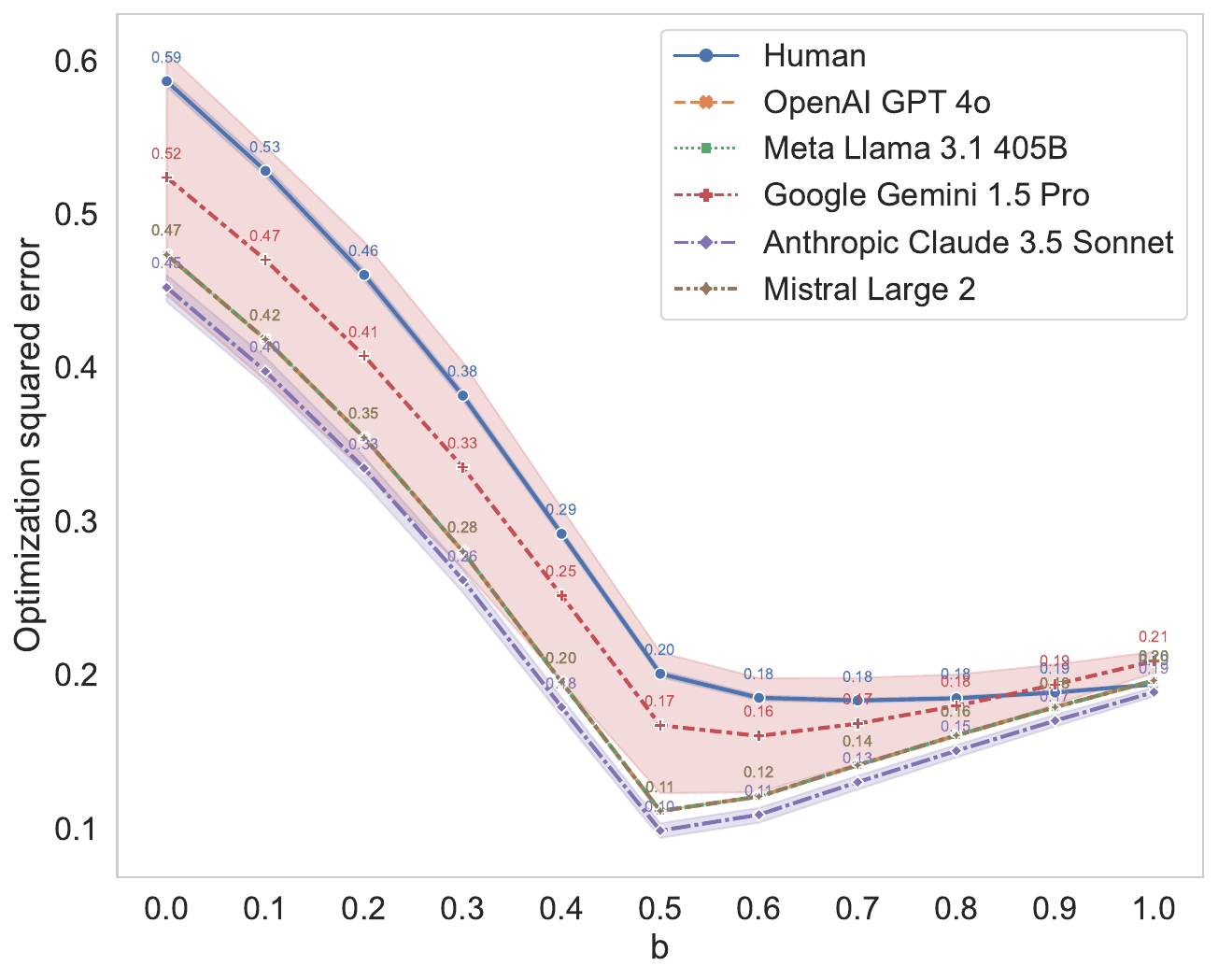}
        \caption{Trust - Investor}
    \end{subfigure}
    \begin{subfigure}[b]{0.24\textwidth}
        \includegraphics[width=\textwidth]{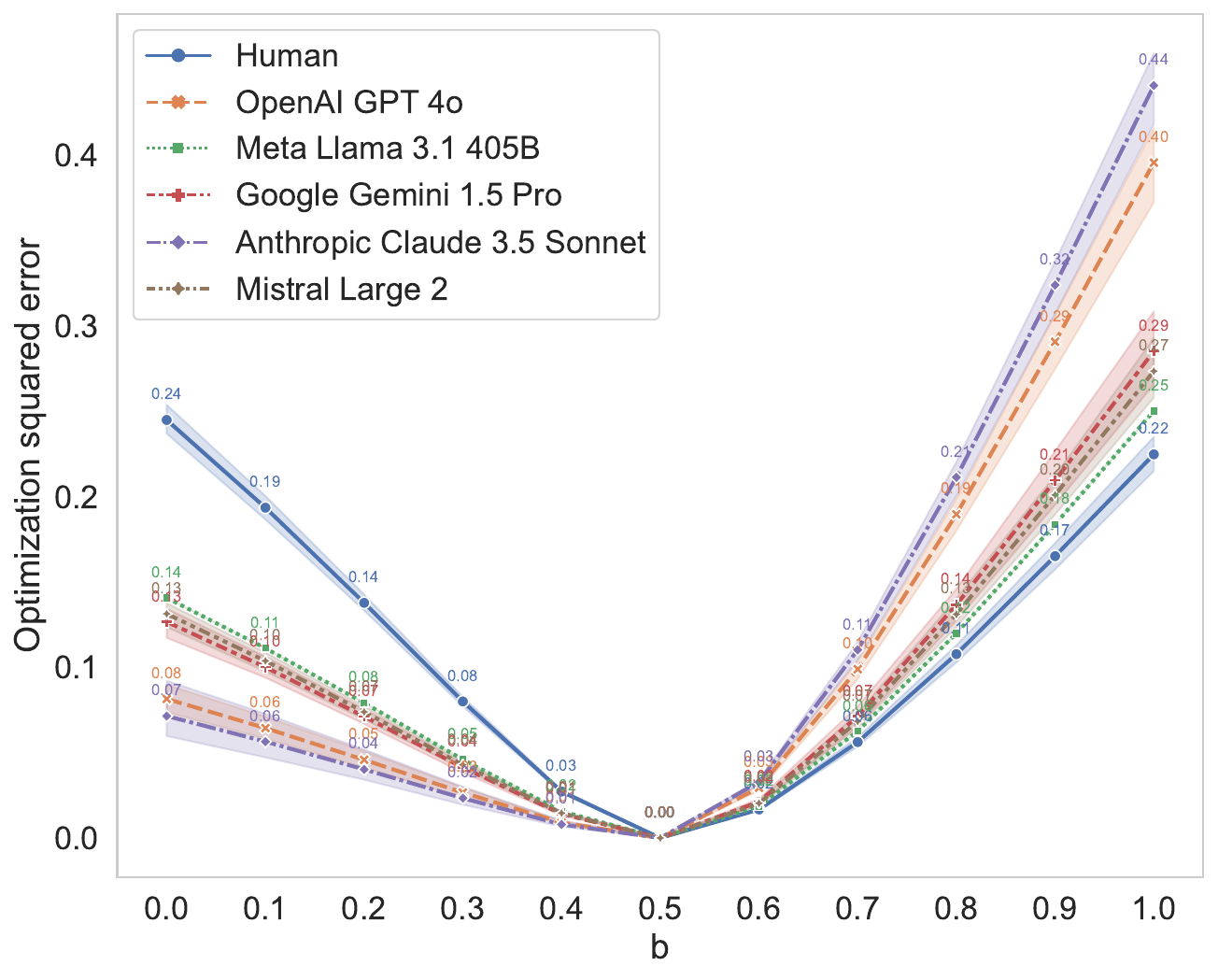}
        \caption{Trust - Banker}
    \end{subfigure}
    \begin{subfigure}[b]{0.24\textwidth}
        \includegraphics[width=\textwidth]{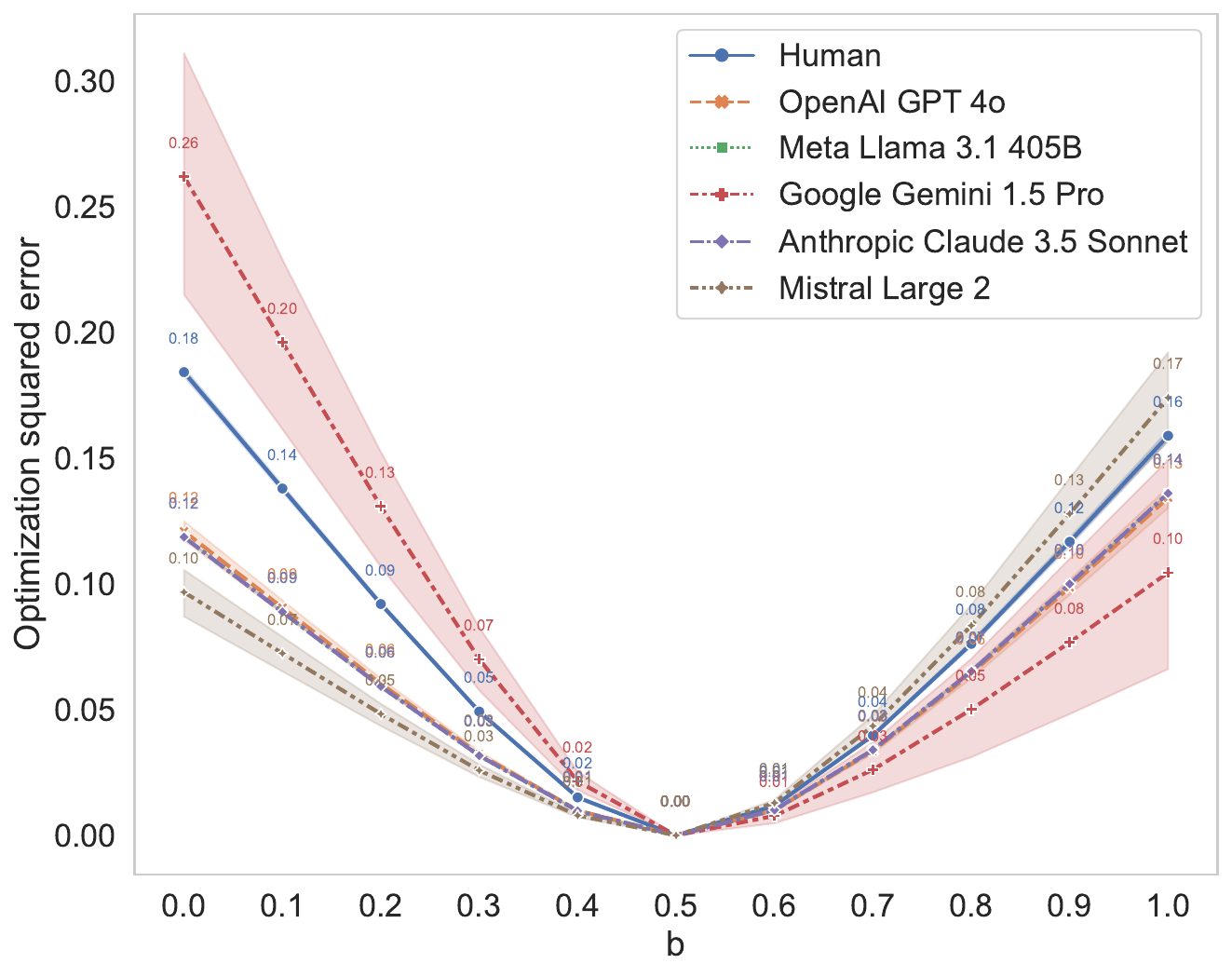}
        \caption{Public - Goods}
    \end{subfigure}
    \begin{subfigure}[b]{0.24\textwidth}
        \includegraphics[width=\textwidth]{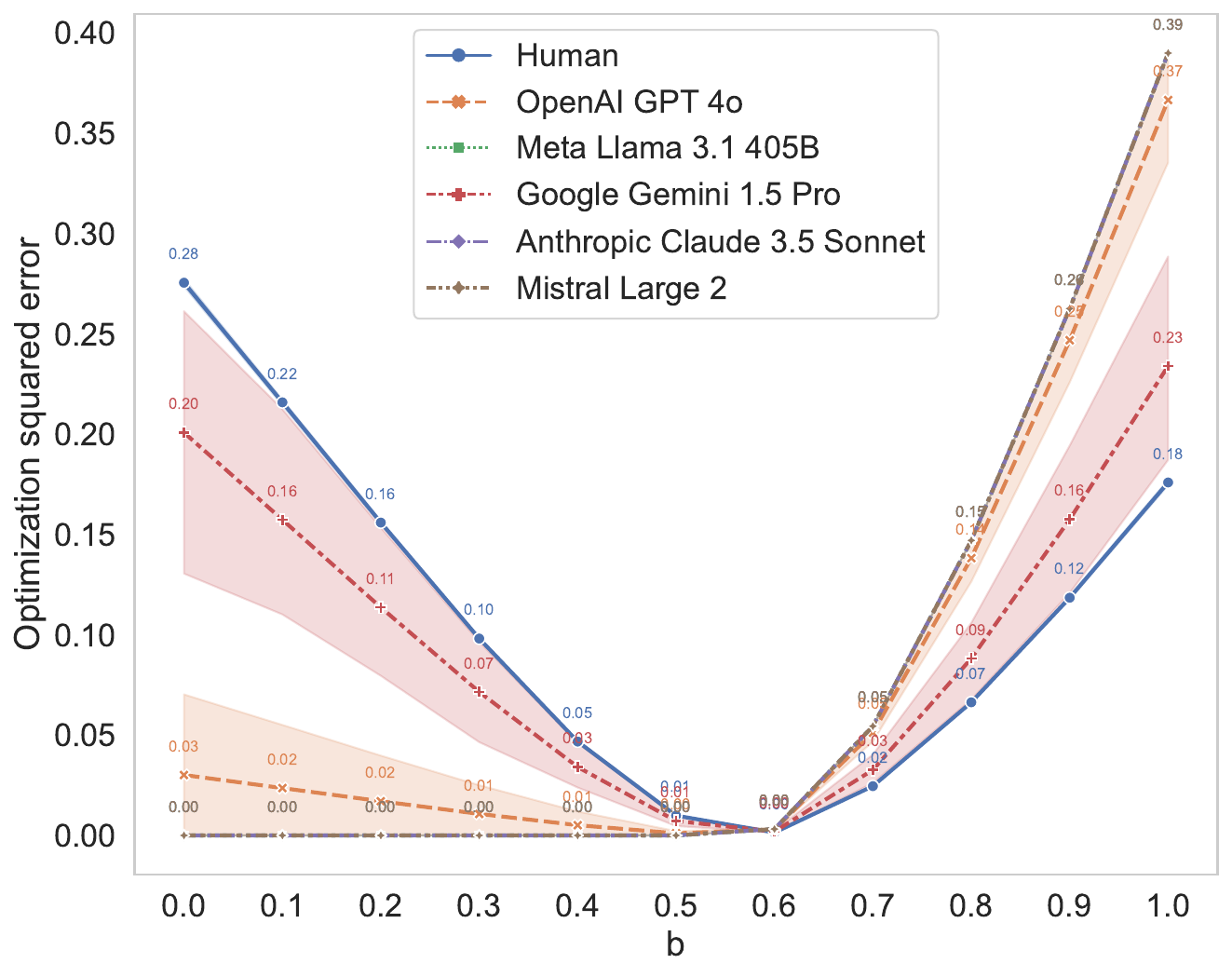}
        \caption{Prisoner's Dilemma}
    \end{subfigure}
    \caption{Mean squared error of the actual distribution of play relative to the best-response payoff, when matched with a partner playing the human distribution. The average is across all games. The errors are reported for each possible b, which is the weight on own vs partner payoff in the utility function (linear blend, with CES specification r = 1). b = 1 is the purely selfish (own) payoff, b = 0 is the purely selfless (partner) payoff, and b = 1/2 is the overall welfare (average) payoff. The values of mean square errors are annotated in the plots.
    }
    \label{fig:mse-diff-games-r1}
\end{figure*}

\begin{table*}[h]
\centering
\caption{Estimation of the weight $b$ by multinomial logit discrete choice analysis. 
}

\label{tab:beta-estimation1}
\resizebox{\textwidth}{!}{
\begin{tabular}{|c|c|c|c|c|c|c|c|}
\hline
\multirow{2}{*}{\textbf{Game}} & \multirow{2}{*}{\textbf{Player}} & \multicolumn{3}{c|}{\textbf{With CES specification $r=1$}} & \multicolumn{3}{c|}{\textbf{With CES specification $r=1/2$}} \\ \cline{3-8} %
& & \textbf{Estimated $b$} & \textbf{Standard error} & \textbf{Confidence interval} & \textbf{Estimated $b$} & \textbf{Standard error} & \textbf{Confidence interval} \\ \hline
\multirow{6}{*}{\begin{tabular}[c]{@{}c@{}}Dictator \end{tabular}} 
& Human & 0.517 & 0.000 & (0.516, 0.517) & 0.658 & 0.000 & (0.657, 0.659) \\

& OpenAI GPT 4o  & 0.500 & 0.002 & (0.495, 0.505) & 0.500 & 0.002 & (0.495, 0.505) \\
& Meta Llama 3.1 405B  & 0.500 & 0.002 & (0.495, 0.505) & 0.500 & 0.002 & (0.495, 0.505) \\
& Google Gemini 1.5 Pro  & 0.498 & 0.002 & (0.493, 0.503) & 0.498 & 0.002 & (0.493, 0.503) \\
& Anthropic Claude 3.5 Sonnet & 0.501 & 0.002 & (0.497, 0.506) & 0.501 & 0.002 & (0.497, 0.506) \\
& Mistral Large 2 & 0.500 & 0.002 & (0.495, 0.505) & 0.500 & 0.002 & (0.495, 0.505) \\

 \hline

\multirow{6}{*}{{\begin{tabular}[c]{@{}c@{}}Ultimatum -\\ Proposer \end{tabular}}} 
 & Human & 1.000 & 0.005 & (0.989, 1.011) & 1.000 & 0.005 & (0.990, 1.01) \\

& OpenAI GPT 4o  & 1.000 & 0.059 & (0.884, 1.116) & 1.000 & 0.059 & (0.884, 1.116) \\
& Meta Llama 3.1 405B & 1.000 & 0.059 & (0.884, 1.116) & 1.000 & 0.059 & (0.884, 1.116) \\
& Google Gemini 1.5 Pro  & \colorbox{cellgreen}{0.503} & 0.006 & (0.492, 0.514) & \colorbox{cellgreen}{0.503} & 0.006 & (0.492, 0.514) \\
& Anthropic Claude 3.5 Sonnet & 1.000 & 0.059 & (0.884, 1.116) & 1.000 & 0.059 & (0.884, 1.116) \\
& Mistral Large 2  & 1.000 & 0.059 & (0.884, 1.116) & 1.000 & 0.059 & (0.884, 1.116) \\

 \hline

 \multirow{6}{*}{{\begin{tabular}[c]{@{}c@{}}Ultimatum -\\ Responder \end{tabular}}} 
 & Human & 1.000 & 0.005 & (0.990, 1.010) & 1.000 & 0.006 & (0.989, 1.011) \\

& OpenAI GPT 4o  & 0.994 & 0.054 & (0.887, 1.100) & 1.000 & 0.060 & (0.883, 1.117) \\
& Meta Llama 3.1 405B  & 1.000 & 0.054 & (0.893, 1.107) & 1.000 & 0.063 & (0.877, 1.123) \\
& Google Gemini 1.5 Pro  & \colorbox{cellgreen}{0.647} & 0.047 & (0.556, 0.739) & \colorbox{cellgreen}{0.635} & 0.049 & (0.539, 0.731) \\
& Anthropic Claude 3.5 Sonnet & 1.000 & 0.054 & (0.893, 1.107) & 1.000 & 0.057 & (0.888, 1.112) \\
& Mistral Large 2  & 1.000 & 0.054 & (0.893, 1.107) & 1.000 & 0.057 & (0.889, 1.111) \\

 \hline

 \multirow{6}{*}{{\begin{tabular}[c]{@{}c@{}}Trust -\\ Investor \end{tabular}}} 
 & Human & 0.535 & 0.000 & (0.535, 0.535) & 0.570 & 0.000 & (0.569, 0.570) \\

 & OpenAI GPT 4o  & 0.530 & 0.002 & (0.525, 0.535) & 0.564 & 0.002 & (0.559, 0.569) \\
& Meta Llama 3.1 405B  & 0.530 & 0.002 & (0.525, 0.535) & 0.564 & 0.002 & (0.559, 0.569) \\
& Google Gemini 1.5 Pro  & 0.528 & 0.003 & (0.523, 0.533) & 0.563 & 0.003 & (0.557, 0.568) \\
& Anthropic Claude 3.5 Sonnet & 0.529 & 0.003 & (0.524, 0.534) & 0.563 & 0.003 & (0.558, 0.568) \\
& Mistral Large 2 & 0.530 & 0.002 & (0.525, 0.535) & 0.564 & 0.002 & (0.559, 0.569) \\

 \hline

\multirow{6}{*}{{\begin{tabular}[c]{@{}c@{}}Trust -\\ Banker \end{tabular}}} 
 & Human & 0.504 & 0.000 & (0.504, 0.505) & 0.475 & 0.001 & (0.473, 0.477) \\

 & OpenAI GPT 4o  & \colorbox{cellgreen}{0.495} & 0.002 & (0.491, 0.498) & \colorbox{cellgreen}{0.381} & 0.005 & (0.371, 0.391) \\
& Meta Llama 3.1 405B  & 0.500 & 0.002 & (0.497, 0.503) & 0.435 & 0.005 & (0.425, 0.445) \\
& Google Gemini 1.5 Pro  & 0.499 & 0.002 & (0.496, 0.502) & 0.422 & 0.005 & (0.412, 0.432) \\
& Anthropic Claude 3.5 Sonnet & \colorbox{cellgreen}{0.494} & 0.002 & (0.490, 0.497) & \colorbox{cellgreen}{0.368} & 0.005 & (0.358, 0.378) \\
& Mistral Large 2  & 0.499 & 0.002 & (0.496, 0.502) & 0.427 & 0.005 & (0.417, 0.437) \\

 \hline

 \multirow{6}{*}{{\begin{tabular}[c]{@{}c@{}}Public -\\Goods \end{tabular}}} 
 & Human & 0.526 & 0.001 & (0.524, 0.528) & 0.518 & 0.001 & (0.516, 0.521) \\

 & OpenAI GPT 4o  & 0.502 & 0.023 & (0.456, 0.548) & 0.489 & 0.025 & (0.439, 0.538) \\
& Meta Llama 3.1 405B  & 0.500 & 0.023 & (0.454, 0.546) & 0.486 & 0.025 & (0.436, 0.536) \\
& Google Gemini 1.5 Pro  & 0.606 & 0.026 & (0.555, 0.657) & 0.621 & 0.029 & (0.563, 0.678) \\
& Anthropic Claude 3.5 Sonnet & \colorbox{cellgreen}{0.468} & 0.024 & (0.422, 0.514) & \colorbox{cellgreen}{0.448} & 0.025 & (0.398, 0.498) \\
& Mistral Large 2  & 0.500 & 0.023 & (0.454, 0.546) & 0.486 & 0.025 & (0.436, 0.536) \\

 \hline

 \multirow{6}{*}{{\begin{tabular}[c]{@{}c@{}}Prisoner's -\\ Dilemma \end{tabular}}} 
 & Human & 0.572 & 0.000 & (0.572, 0.572) & 0.563 & 0.000 & (0.563, 0.563) \\

 & OpenAI GPT 4o & 0.567 & 0.001 & (0.566, 0.569) & 0.559 & 0.001 & (0.557, 0.561) \\
& Meta Llama 3.1 405B & 0.566 & 0.001 & (0.563, 0.569) & 0.557 & 0.001 & (0.555, 0.560) \\
& Google Gemini 1.5 Pro  & 0.571 & 0.000 & (0.570, 0.572) & 0.562 & 0.000 & (0.562, 0.563) \\
& Anthropic Claude 3.5 Sonnet  & 0.566 & 0.001 & (0.563, 0.569) & 0.557 & 0.001 & (0.555, 0.560) \\
& Mistral Large 2  & 0.566 & 0.001 & (0.563, 0.569) & 0.557 & 0.001 & (0.555, 0.560) \\

 \hline

\end{tabular}
}
\vspace{5pt}

$^*:$ To be comparable, the Trust-Banker calculations are done assuming that the original investment is \$50.  \\ $^\dagger:$ The Prisoner's Dilemma reports the estimation results in the first round of the game.
\end{table*}

\begin{figure*}[h]
    \centering
    \begin{subfigure}[b]{0.24\textwidth}
        \includegraphics[width=\textwidth]{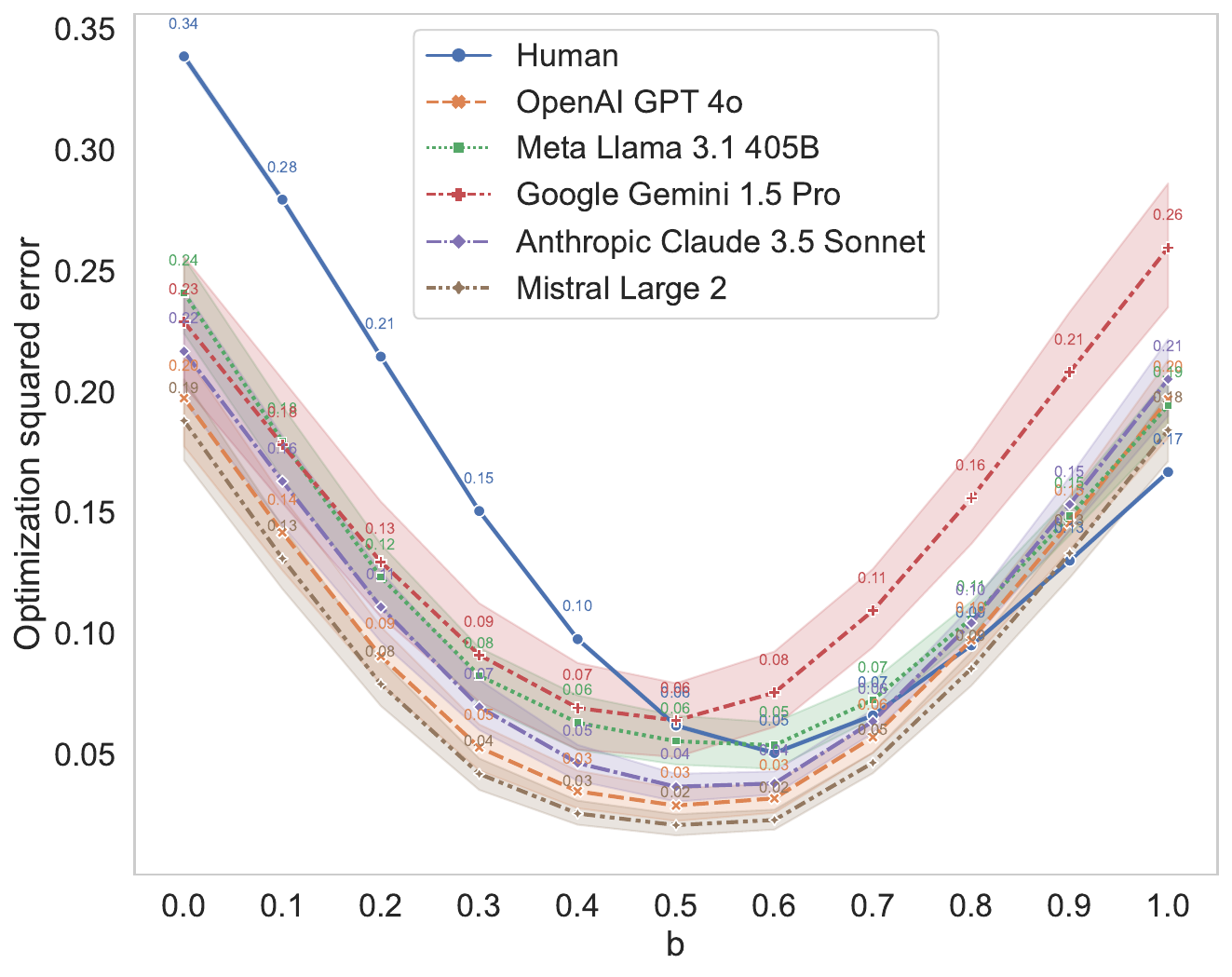}
        \caption{Average}
    \end{subfigure}
    \begin{subfigure}[b]{0.24\textwidth}
        \includegraphics[width=\textwidth]{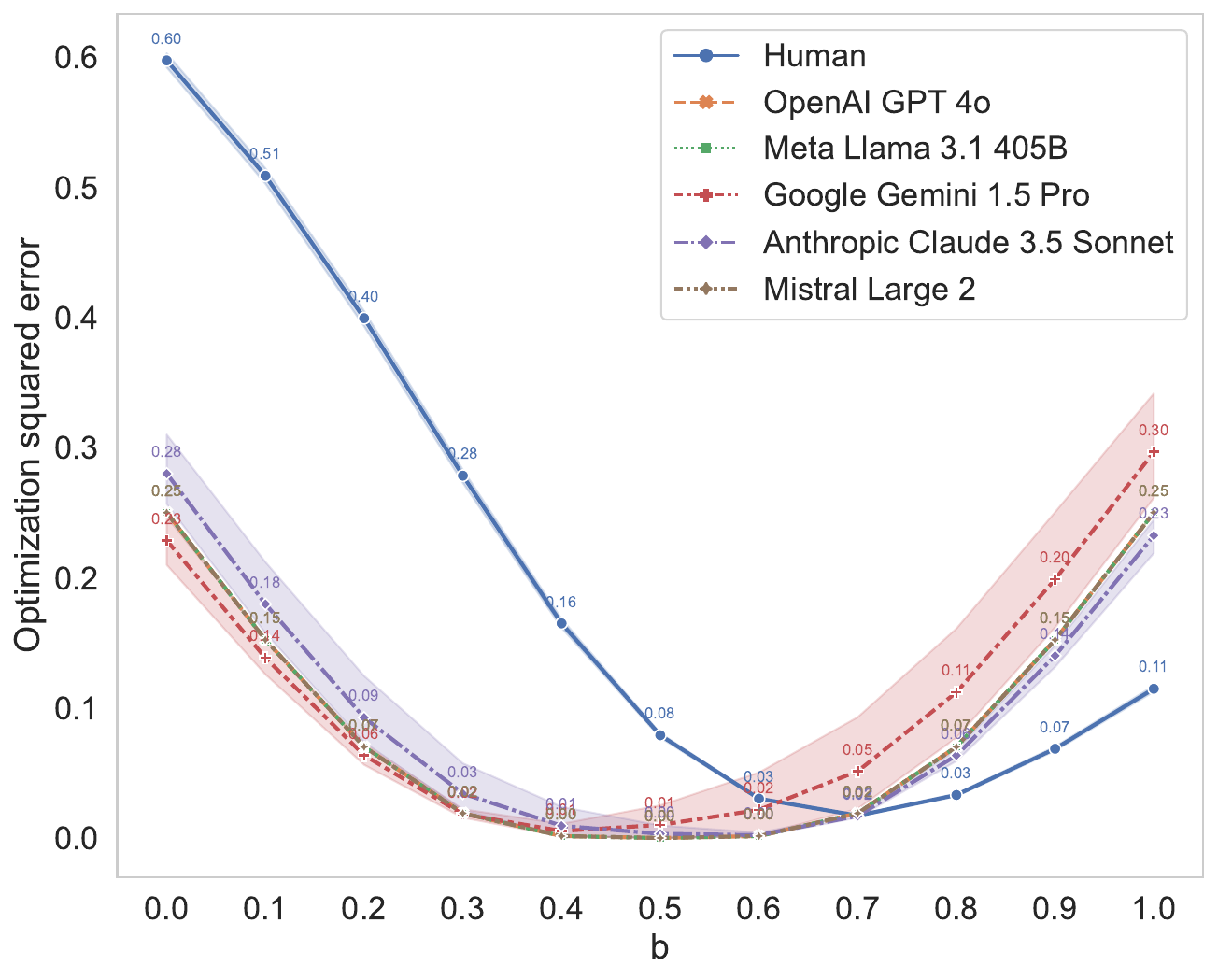}
        \caption{Dictator}
    \end{subfigure}
    \begin{subfigure}[b]{0.24\textwidth}
        \includegraphics[width=\textwidth]{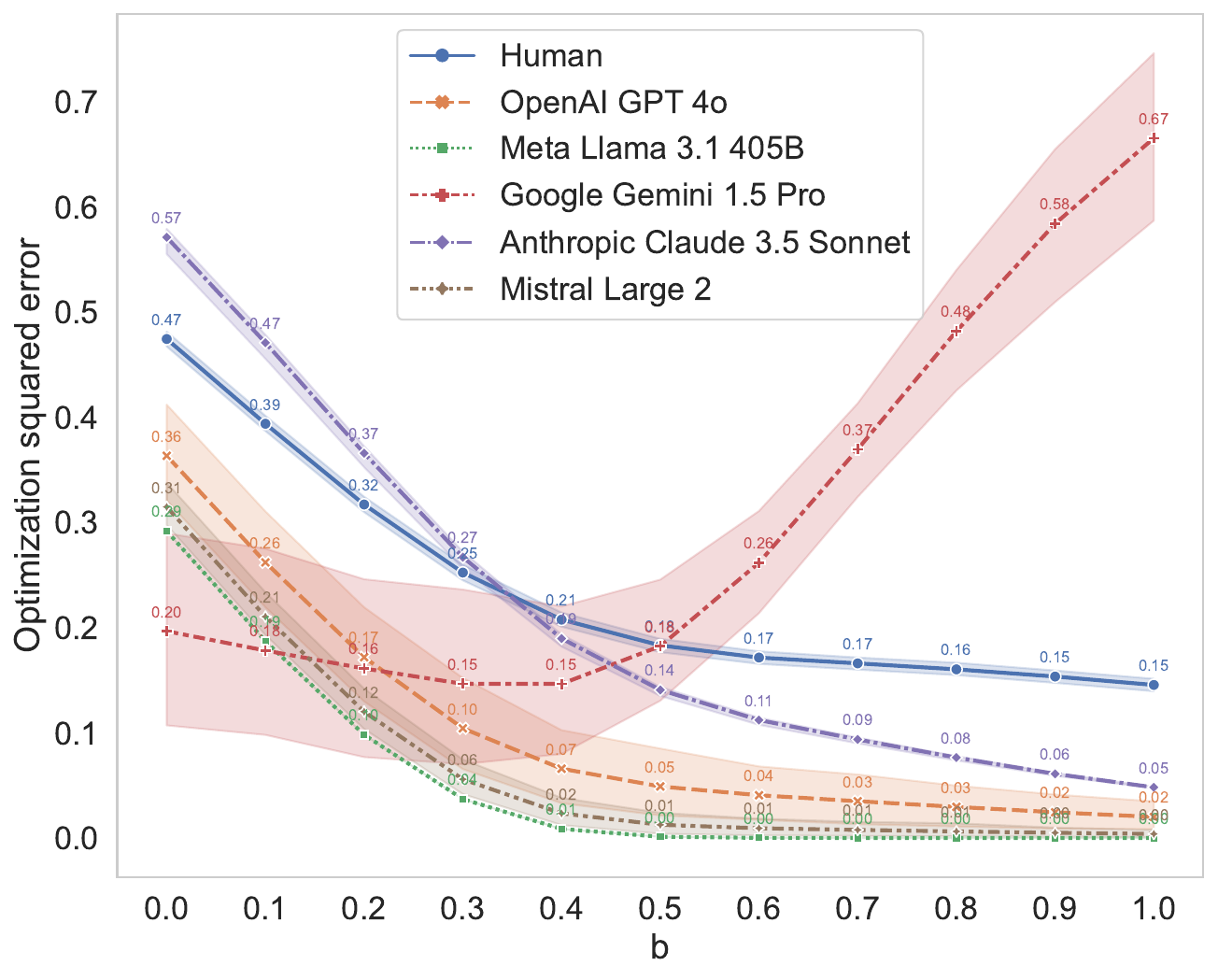}
        \caption{Ultimatum - Proposer}
    \end{subfigure}
    \begin{subfigure}[b]{0.24\textwidth}
        \includegraphics[width=\textwidth]{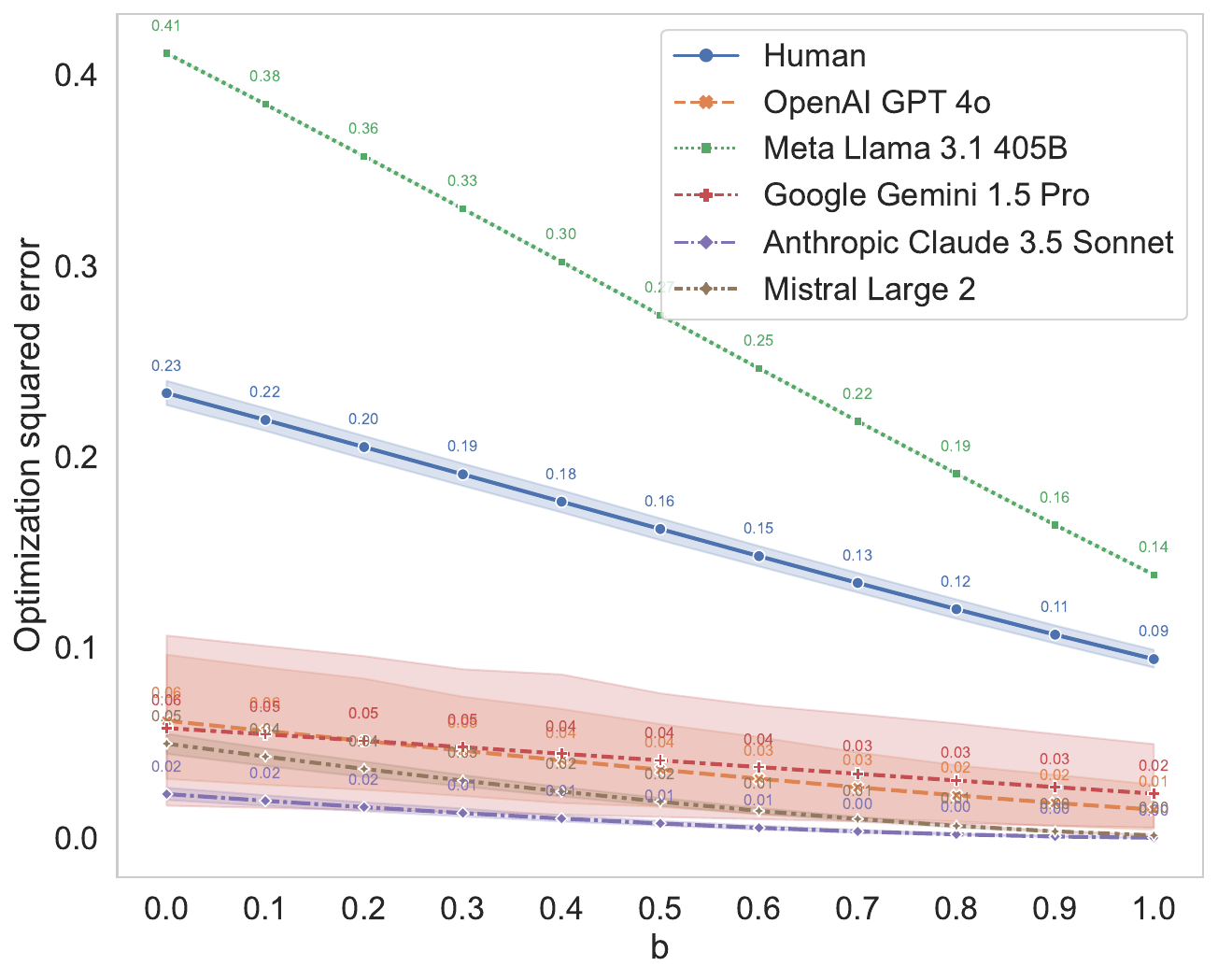}
        \caption{Ultimatum - Responder}
    \end{subfigure}
    
    \begin{subfigure}[b]{0.24\textwidth}
        \includegraphics[width=\textwidth]{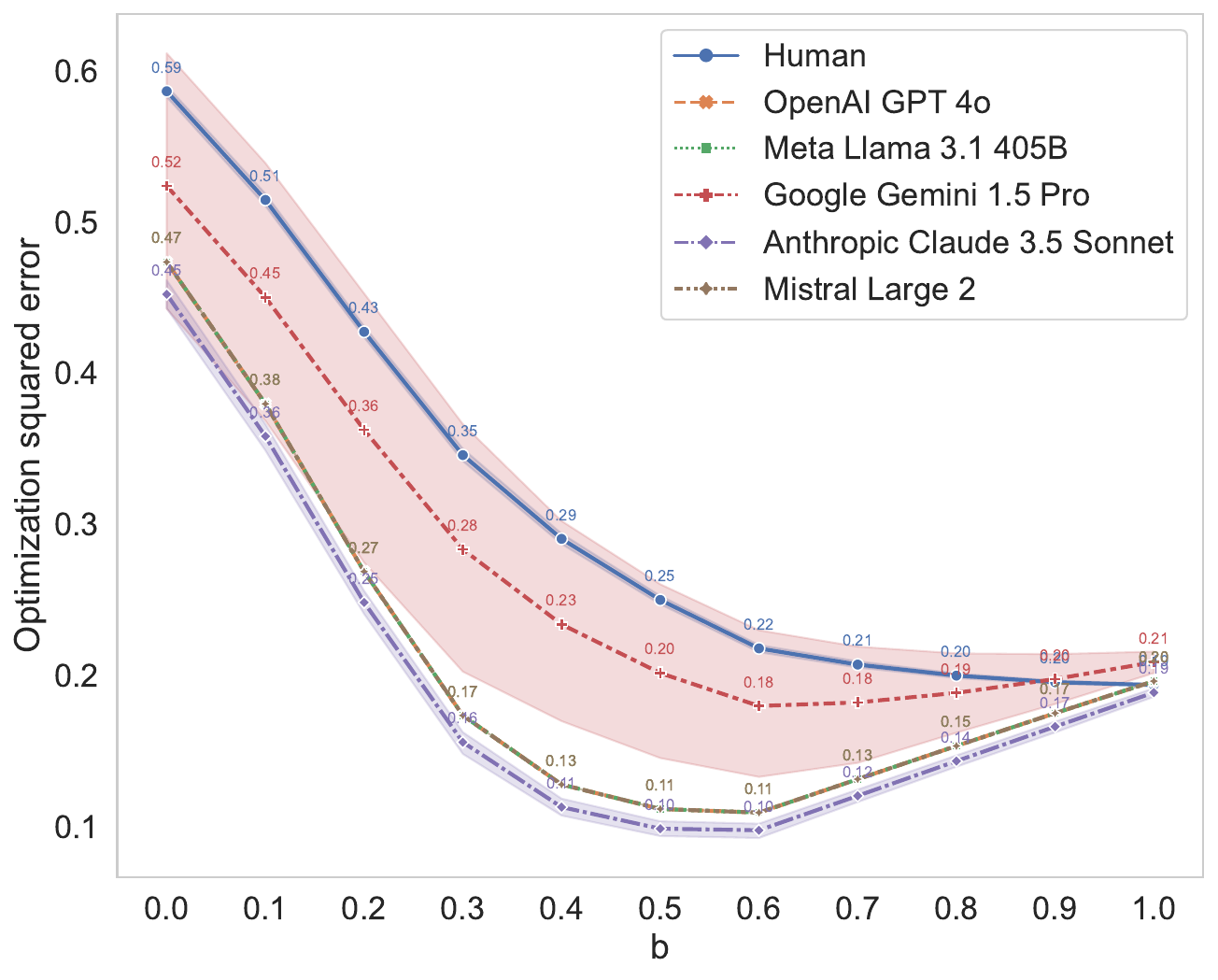}
        \caption{Trust - Investor}
    \end{subfigure}
    \begin{subfigure}[b]{0.24\textwidth}
        \includegraphics[width=\textwidth]{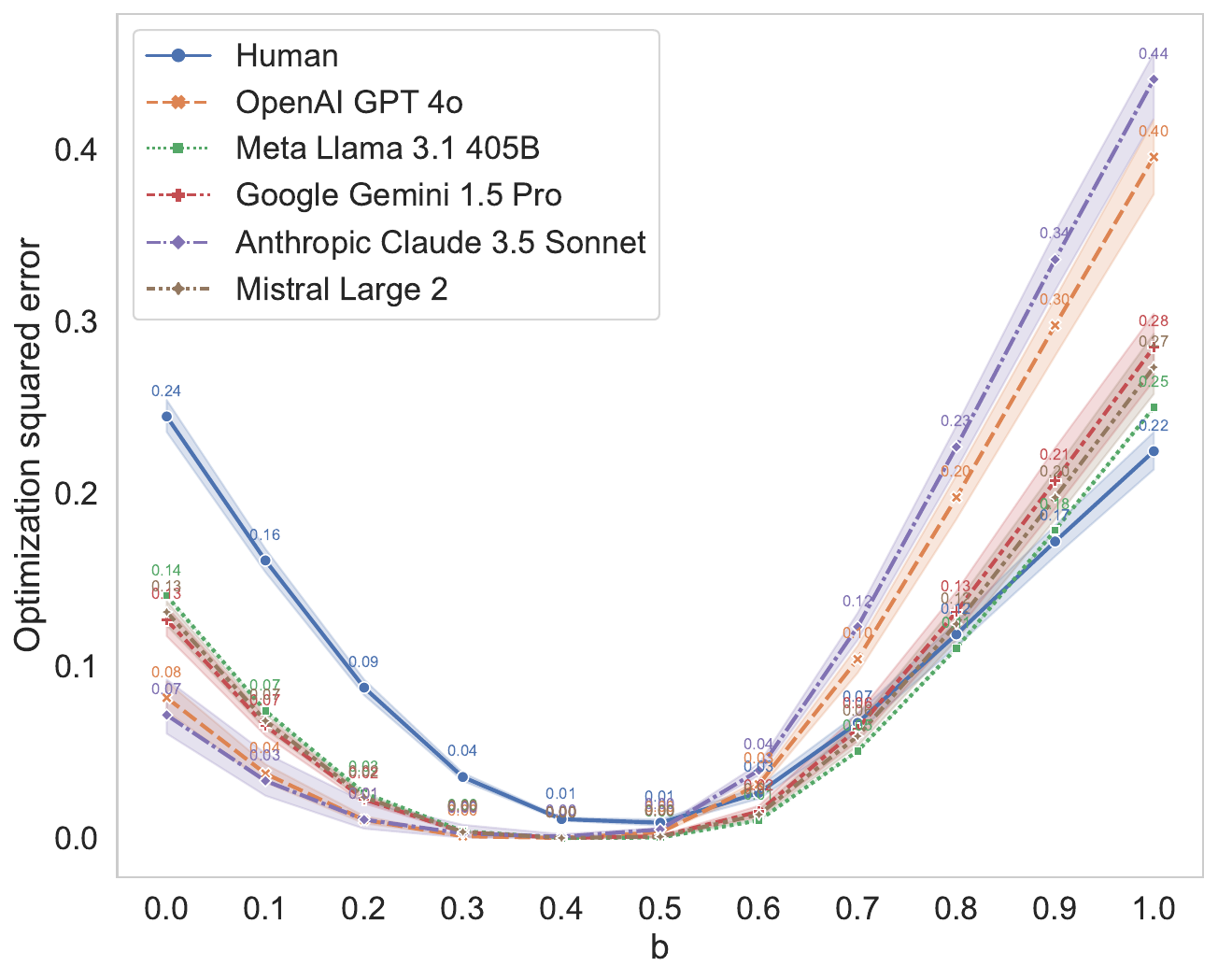}
        \caption{Trust - Banker}
    \end{subfigure}
    \begin{subfigure}[b]{0.24\textwidth}
        \includegraphics[width=\textwidth]{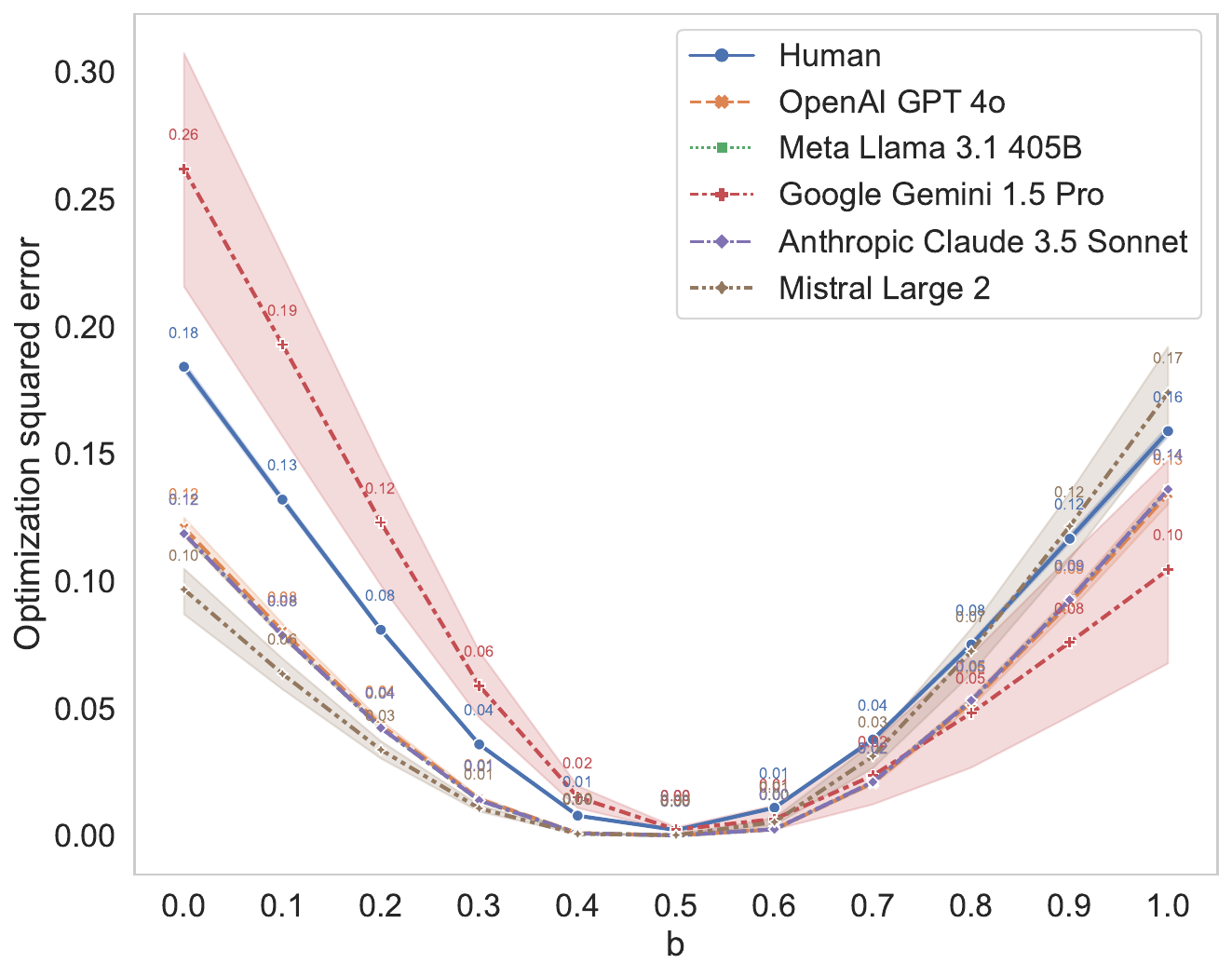}
        \caption{Public - Goods}
    \end{subfigure}
    \begin{subfigure}[b]{0.24\textwidth}
        \includegraphics[width=\textwidth]{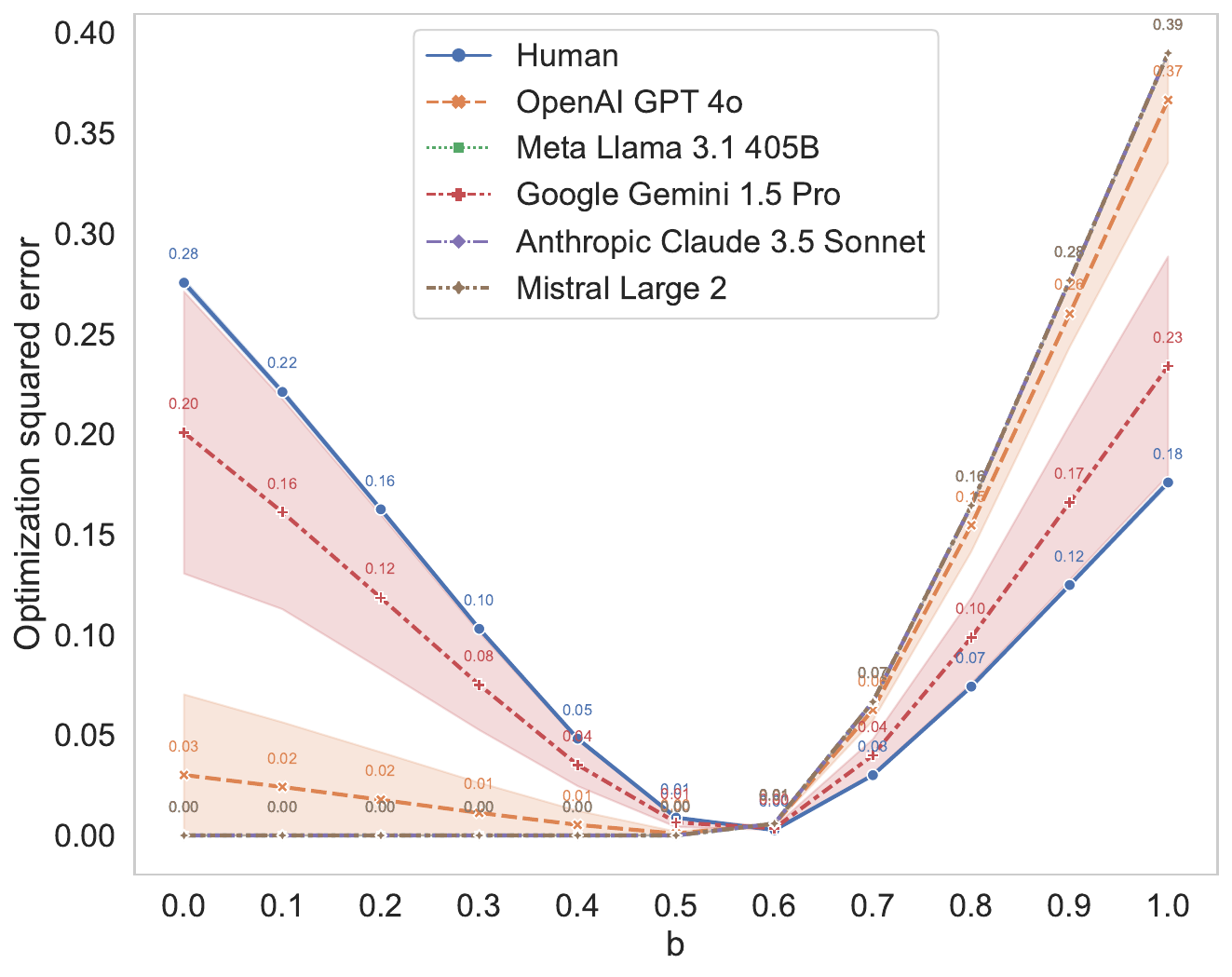}
        \caption{Prisoner's Dilemma}
    \end{subfigure}
    \caption{Mean squared error of the actual distribution of play relative to the best-response payoff, when matched with a partner playing the human distribution. The average is across all games. The errors are reported for each possible b, which is the weight on own vs partner payoff in the utility function (non-linear blend, with CES specification
r = 1/2). b = 1 is the purely selfish (own) payoff, and b = 0 is the purely selfless (partner) payoff. The values of mean square errors are annotated in the plots.
    }
    \label{fig:mse-diff-games-r0.5}
\end{figure*}

\begin{table*}[h]
    \centering
    \begin{tabular}{|l|l|c|c|}
        \hline
        \multirow{2}{*}{\textbf{AI Chatbot family}} & \multirow{2}{*}{\textbf{Player}} & \multicolumn{2}{c|}{\textbf{Inconsistency}}  \\ \cline{3-4}
        & & $r=1.0$ & $r=0.5$ \\ \hline
	&	\phantom{0}Human players	&	0.114	&	0.122	\\ \hline
OpenAI GPT	&	\texttt{	gpt4-4o-2024-05-31	}	&	0.115	&	0.107	\\
	&	\texttt{	gpt4-4o-mini-2024-07-18	}	&	0.148	&	0.140	\\
	&	\texttt{	gpt-4-0125-preview	}	&	0.124	&	0.115	\\
	&	\texttt{	gpt-4-0613	}	&	\textbf{0.096}	&	\textbf{0.090}	\\
	&	\texttt{	gpt-3.5-turbo-0125	}	&	0.160	&	0.154	\\
	&	\texttt{	gpt-3.5-turbo-0613	}	&	0.168	&	0.185	\\ \hline
Meta Llama	&	\texttt{	llama-3.1-405B-instruct	}	&	0.125	&	0.125	\\
	&	\texttt{	llama-3-70b-chat	}	&	0.205	&	0.201	\\
	&	\texttt{	llama-3-8b-chat	}	&	0.183	&	0.182	\\ \hline
Google Gemini	&	\texttt{	gemini-1.5-pro-latest	}	&	0.118	&	0.139	\\
	&	\texttt{	gemini-1.0-pro-001	}	&	0.146	&	0.152	\\ \hline
Anthropic Claude	&	\texttt{	claude-3-5-sonnet-20240620	}	&	0.143	&	0.133	\\
	&	\texttt{	claude-3-opus-20240229	}	&	0.115	&	0.104	\\
	&	\texttt{	claude-3-sonnet-20240229	}	&	0.117	&	0.111	\\
	&	\texttt{	claude-3-haiku-20240307	}	&	0.123	&	0.115	\\ \hline
Mistral	&	\texttt{	mistral-large-2407	}	&	0.108	&	0.100	\\
	&	\texttt{	mistral-large-2402	}	&	0.134	&	0.128	\\
    \hline
    \end{tabular}
    \caption{Behavior inconsistency across games of AI chatbots. The inconsistency is estimated by the mean absolute error of payoff curves. }
    \label{tab:inconsistency-all}
\end{table*}

\begin{figure*}[h]
    \centering
    \includegraphics[width=\textwidth]{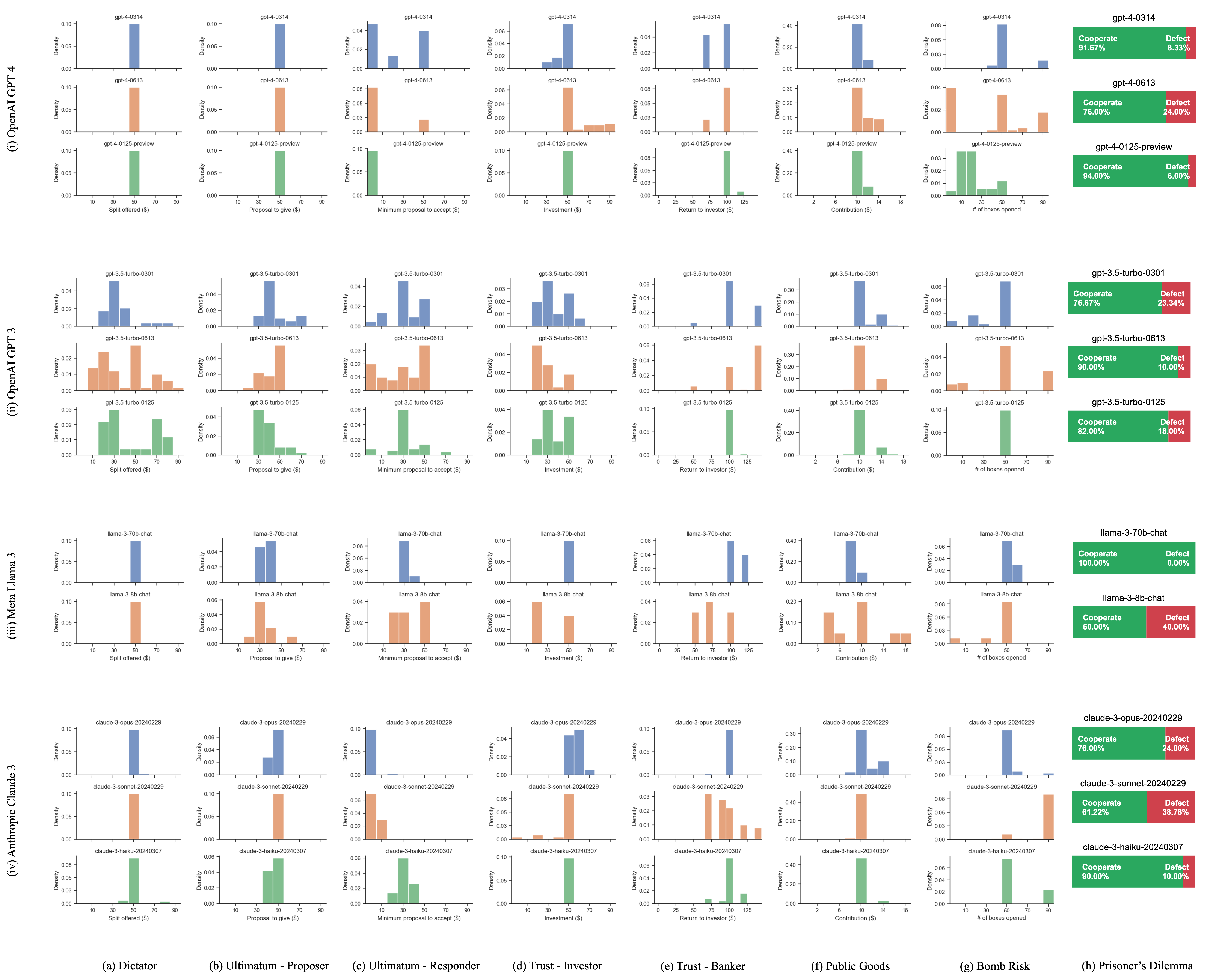}
    \caption{Distributions of AI chatbot behaviors in behavioral economics games. }
    \label{fig:variants}
\end{figure*}

\end{document}